\documentclass[runningheads]{llncs}
\usepackage{graphicx}
\usepackage{tikz}
\usepackage{comment}
\usepackage{graphicx}
\usepackage{amsmath,amssymb} % define this before the line numbering.
\usepackage{color}

\usepackage{booktabs}
\usepackage{array}
\newsavebox\dummy
\newcolumntype{H}{>{\begin{lrbox}{\dummy}}c<{\end{lrbox}}@{}}

\usepackage{xcolor}
\definecolor{maroon}{rgb}{0.5, 0.0, 0.0}
\definecolor{ao}{rgb}{0.0, 0.5, 0.0}
\usepackage[labelsep=period]{caption}
\usepackage{subcaption}
\usepackage{algorithm}
\usepackage[noend]{algorithmic}
\usepackage{amsmath}
\usepackage{multirow}
\DeclareMathOperator{\LPIPS}{LPIPS} 
\DeclareMathOperator{\KL}{KL} 

\definecolor{NVblue}{rgb}{0.07,0.12,0.83}
\definecolor{BUred}{rgb}{0.8,0.0,0.0}
\usepackage[accsupp]{axessibility}  % Improves PDF readability for those with disabilities.

\usepackage[pagebackref=true,breaklinks=true,colorlinks=true,citecolor=NVblue,linkcolor=BUred,bookmarks=false]{hyperref}
 \usepackage{amssymb}
 \usepackage{orcidlink}
\begin{document}

\pagestyle{headings}
\mainmatter
\def\ECCVSubNumber{7573}  % Insert your submission number here
\newcounter{myequation}
\makeatletter
\@addtoreset{equation}{myequation}
\makeatother

\newcounter{myalgorithm}
\makeatletter
\@addtoreset{algorithm}{myalgorithm}
\makeatother

\newcounter{mytable}
\makeatletter
\@addtoreset{table}{mytable}
\makeatother

\newcounter{mysection}
\makeatletter
\@addtoreset{section}{mysection}
\makeatother

\newcounter{myfigure}
\makeatletter
\@addtoreset{figure}{myfigure}
\makeatother
\title{Scaling Adversarial Training to Large Perturbation Bounds} % Replace with your title

\begin{comment}
\titlerunning{ECCV-22 submission ID \ECCVSubNumber} 
\authorrunning{ECCV-22 submission ID \ECCVSubNumber} 
\author{Anonymous ECCV submission}
\institute{Paper ID \ECCVSubNumber}
\end{comment}
%******************

% \begin{comment}
\titlerunning{Scaling Adversarial Training to Large Perturbation Bounds}
\author{
  Sravanti Addepalli$^\dagger$ \thanks{Equal Contribution. \\ Correspondence to Sravanti Addepalli $<$sravantia@iisc.ac.in$>$, Samyak Jain $<$samyakjain.cse18@itbhu.ac.in$>$ \\ $^\ddagger$ Work done during internship at Video Analytics Lab, Indian Institute of Science.} \orcidlink{0000-0001-7238-4603} \qquad Samyak Jain$^\dagger$ $^\diamond$ $^\ddagger$ $^{\star}$ \orcidlink{0000-0003-3785-4782} \qquad  Gaurang Sriramanan$^\dagger$\orcidlink{0000-0001-7008-2417} \qquad R.Venkatesh Babu$^\dagger$ \orcidlink{0000-0002-1926-1804}
  }

%\author{Sravanti Addepalli\thanks{Equal contribution. \newline Correspondence to: Sravanti Addepalli $<$sravantia@iisc.ac.in$>$, Samyak Jain $<$samyakjain.cse18@itbhu.ac.in$>$ \\ $^\ddagger$ Work done during internship at Video Analytics Lab, Indian Institute of Science.} \orcidlink{0000-0001-7238-4603} \and
%Samyak Jain$^{\star}$$^\diamond$ $^\ddagger$ \footnotemark[1] \orcidlink{0000-0003-3785-4782} \and
%Gaurang Sriramanan \orcidlink{0000-0001-7008-2417} \and
%R.Venkatesh Babu \orcidlink{0000-0002-1926-1804}}

\authorrunning{S. Addepalli et al.}
\institute{$^\dagger$ Video Analytics Lab, Indian Institute of Science, Bangalore  \\ $^\diamond$ Indian Institute of Technology (BHU) Varanasi \hfill}
% \end{comment}
%******************
\maketitle

\begin{abstract}
The vulnerability of Deep Neural Networks to Adversarial Attacks has fuelled research towards building robust models. While most Adversarial Training algorithms aim at defending attacks constrained within low magnitude Lp norm bounds, real-world adversaries are not limited by such constraints. In this work, we aim to achieve adversarial robustness within larger bounds, against perturbations that may be perceptible, but do not change human (or Oracle) prediction. The presence of images that flip Oracle predictions and those that do not makes this a challenging setting for adversarial robustness. We discuss the ideal goals of an adversarial defense algorithm beyond perceptual limits, and further highlight the shortcomings of naively extending existing training algorithms to higher perturbation bounds. In order to overcome these shortcomings, we propose a novel defense, Oracle-Aligned Adversarial Training (OA-AT), to align the predictions of the network with that of an Oracle during adversarial training. The proposed approach achieves state-of-the-art performance at large epsilon bounds (such as an L-inf bound of 16/255 on CIFAR-10) while outperforming existing defenses (AWP, TRADES, PGD-AT) at standard bounds (8/255) as well.
\end{abstract}

\section{Introduction}

Deep Neural Networks are known to be vulnerable to Adversarial Attacks, which are perturbations crafted with an intention to fool the network \cite{intriguing-iclr-2014}. With the rapid increase in deployment of Deep Learning algorithms in various critical applications such as autonomous navigation, it is becoming increasingly crucial to improve the Adversarial robustness of these models. 
In a classification setting, Adversarial attacks can flip the prediction of a network to even unrelated classes, while causing no change in a human's prediction (Oracle label).

\begin{figure}[t]
\centering
        \includegraphics[width=0.9\linewidth]{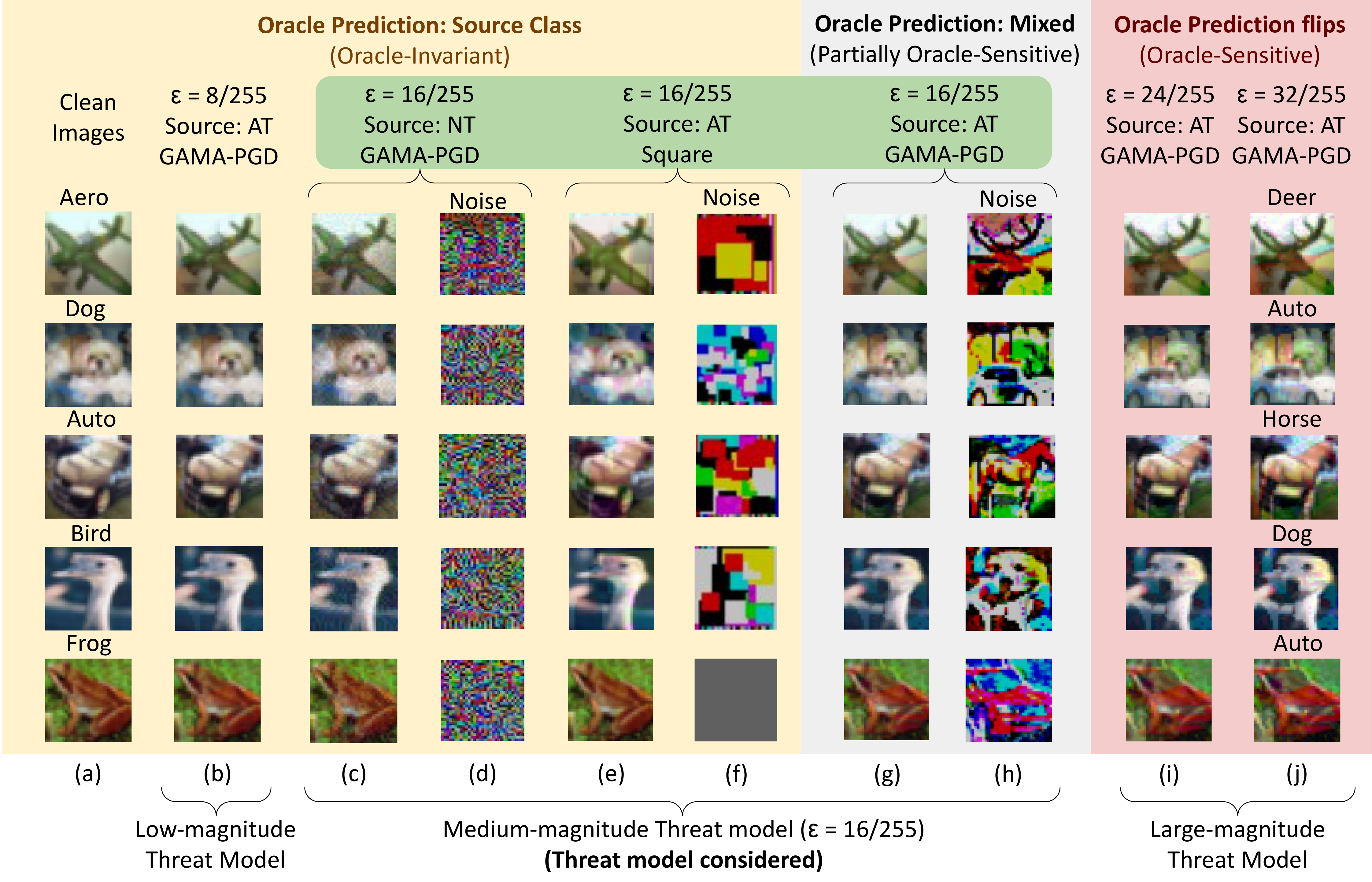}
        \caption{\textbf{Perturbations within different threat models:} Adversarial images (b, c, e, g, i, j) and perturbations (d, f, h) along with the corresponding clean image (a) for various $\ell_\infty$ norm bounds on CIFAR-10. Attacks are generated from an Adversarially Trained model (AT) or a Normally Trained model (NT) using the gradient-based attack GAMA-PGD \cite{sriramanan2020gama} or the Random-search based attack Square \cite{andriushchenko2019square}. The medium-magnitude threat model is challenging since it consists of attacks which are Oracle-Invariant and partially Oracle-Sensitive.}
        %\vspace{-0.3cm}
        \label{fig:lp_pert}
\end{figure}

The definition of adversarial attacks involves the prediction of an Oracle, making it challenging to formalize threat models for the training and verification of adversarial defenses. 
The widely used convention that overcomes this challenge is the $\ell_p$ norm based threat model with low-magnitude bounds to ensure imperceptibility \cite{goodfellow2014explaining,carlini2019evaluating}. For example, attacks constrained within an $\ell_\infty$ norm of $8/255$ on the CIFAR-$10$ dataset are imperceptible to the human eye as shown in Fig.\ref{fig:lp_pert}(b), ensuring that the Oracle label is unchanged. The goal of Adversarial Training within such a threat model is to ensure that the prediction of the model is consistent within the considered perturbation radius $\varepsilon$, and matches the label associated with the unperturbed image. 

\begin{table*}
\begin{minipage}{0.48\linewidth}
\caption{\textbf{CIFAR-10: Standard Adversarial Training using Large-$\varepsilon$:} Performance ($\%$) of various existing Defenses trained using $\varepsilon=8/255$ or $16/255$ against attacks bound within $\varepsilon=8/255$ and $16/255$.  A large drop in clean accuracy is observed with existing approaches \cite{zhang2019theoretically,wu2020adversarial,madry-iclr-2018,zhang2020attacks} when trained using perturbations with $\varepsilon=16/255$.} 
\setlength\tabcolsep{3pt}
\resizebox{1.0\linewidth}{!}{
\label{table:cifar10_16_main}
\begin{tabular}{lccHHccHHHHHHcc}
\toprule
\textbf{Method}              & \textbf{\begin{tabular}[c]{@{}c@{}}Attack $\varepsilon$ \\ \small{(Training)}\end{tabular}} & \textbf{\begin{tabular}[c]{@{}c@{}}Clean \\ Acc\end{tabular}} & \textbf{\begin{tabular}[c]{@{}c@{}}FGSM (BB)\\ (8/255)\end{tabular}} & \textbf{\begin{tabular}[c]{@{}c@{}}R-FGSM \\ (8/255)\end{tabular}} & \textbf{\begin{tabular}[c]{@{}c@{}}GAMA \\ (8/255)\end{tabular}} & \textbf{\begin{tabular}[c]{@{}c@{}}AA \\ (8/255)\end{tabular}} & \textbf{\begin{tabular}[c]{@{}c@{}}FGSM (BB)\\ (12/255)\end{tabular}} & \textbf{\begin{tabular}[c]{@{}c@{}}R-FGSM \\ (12/255)\end{tabular}} & \textbf{\begin{tabular}[c]{@{}c@{}}GAMA \\ (12/255)\end{tabular}} & \textbf{\begin{tabular}[c]{@{}c@{}}Square \\ (12/255)\end{tabular}} & \textbf{\begin{tabular}[c]{@{}c@{}}FGSM (BB)\\ (16/255)\end{tabular}} & \textbf{\begin{tabular}[c]{@{}c@{}}R-FGSM \\ (16/255)\end{tabular}} & \textbf{\begin{tabular}[c]{@{}c@{}}GAMA \\ (16/255)\end{tabular}} & \textbf{\begin{tabular}[c]{@{}c@{}}Square \\ (16/255)\end{tabular}} \\
\midrule
TRADES                       & 8/255                                                              & 80.53          & 78.58                                                                & 63.69                                                              & 49.63                                                                & 49.42                                                          & 77.20                                                                 & 55.48                                                               & 33.32                                                                 & 40.94                                                               & 75.05                                                                 & \textbf{47.92}                                                               & 19.27                                                                 & 27.82                                                               \\
TRADES                       & 16/255                                                             & \textcolor{maroon}{75.30}          & 73.26                                                                & 53.10                                                              & 35.64                                                                & 35.12                                                          & 72.13                                                                 & 44.27                                                               & 20.24                                                                 & 30.11                                                               & 70.76                                                                 & 36.99                                                               & 10.10                                                                 & 18.87                                                               \\
AWP                          & 8/255                                                              & 80.47          & 78.22                                                                & 63.32                                                              & \textbf{50.06}                                                                & \textbf{49.87}                                                          & 76.88                                                                 & 54.61                                                               & 33.47                                                                 & 41.05                                                               & 74.42                                                                 & 46.16                                                               & 19.66                                                                 & 28.51                                                               \\
AWP                          & 16/255                                                             & \textcolor{maroon}{71.63}          & 69.71                                                                & 54.53                                                              & 40.85                                                                & 40.55                                                          & 68.65                                                                 & 47.13                                                               & 27.06                                                                 & 34.42                                                               & 67.42                                                                 & 40.89                                                               & 15.92                                                                 & 24.16                                                               \\

PGD-AT                       & 8/255                                                              & 81.12          & 78.94                                                                & 63.48                                                              & 49.03                                                                & 48.58                                                          & 77.19                                                                 & 54.42                                                               & 30.84                                                                 & 40.82                                                               & 74.37                                                                 & 46.28                                                               & 15.77                                                                 & 26.47                                                               \\
PGD-AT                       & 16/255                                                             & \textcolor{maroon}{64.93}          & 63.65                                                                & 55.47                                                              & 46.66                                                                & 46.21                                                          & 62.81                                                                 & 51.05                                                               & 36.95                                                                 & 40.53                                                               & 61.70                                                                 & 46.40                                                               & \textbf{26.73}                                                                 & \textbf{32.25}                                                               \\
FAT                          & 8/255                                                              & \textbf{84.36} & \textbf{82.20}                                                                & \textbf{64.06}                                                              & 48.41                                                                & 48.14                                                          & 80.32                                                                 & 55.41                                                               & 29.39                                                                 & 39.48                                                               & \textbf{78.13}                                                                 & 47.50                                                               & 15.18                                                                 & 25.07                                                               \\
FAT                          & 16/255                                                             & \textcolor{maroon}{75.27}          & 73.44                                                                & 60.25                                                              & 47.68                                                                & 47.34                                                          & 72.22                                                                 & 53.17                                                               & 34.31                                                                 & 39.79                                                               & 70.73                                                                 & 46.88                                                               & 22.93                                                                 & 29.47                                                               \\
\bottomrule
\end{tabular}}
\end{minipage}
\hfill
% \vspace{-0.3cm}
	\begin{minipage}{0.48\linewidth}
		\centering
		\includegraphics[width=\linewidth]{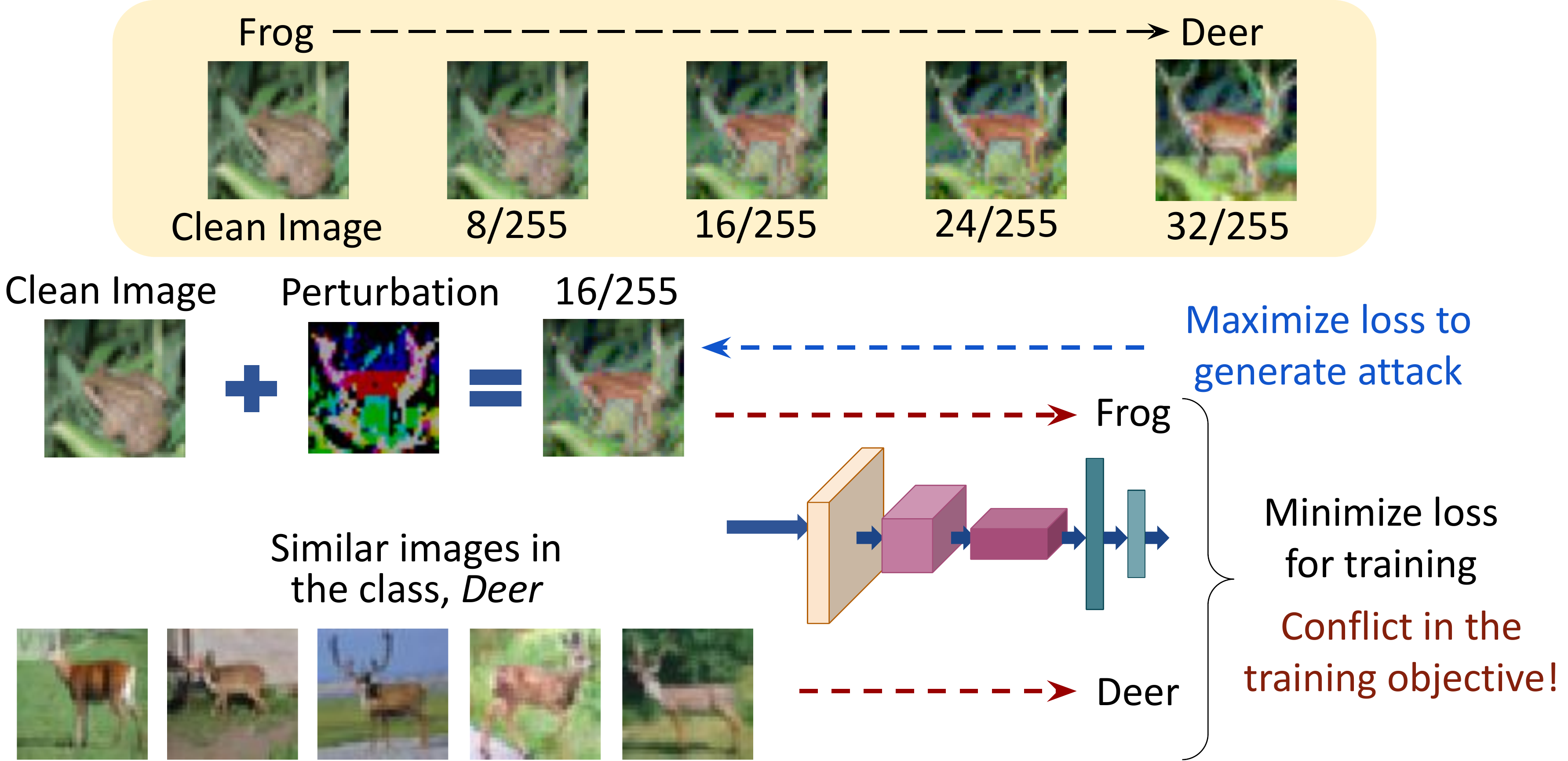}
	    \vspace{1pt}
		\captionof{figure}{\textbf{Issues with Standard Adversarial Training at Large-$\varepsilon$:} An adversarial example generated from the original image of a frog looks partially like a deer at an $\ell_\infty$ bound of $16/255$, but is trained to predict the true label, Frog. This induces a conflicting objective, leading to a large drop in clean accuracy.}
		\label{fig:issue with AT} 
	\end{minipage}
	
% \vspace{-0.7cm}
\end{table*}

While low-magnitude $\ell_p$ norm based threat models form a crucial subset of the widely accepted definition of adversarial attacks \cite{goodfellow_papernot}, they are not sufficient, as there exist valid attacks at higher perturbation bounds as well, as shown in Fig.\ref{fig:lp_pert}(c) and (e). However, the challenge at large perturbation bounds is the existence of attacks that can flip Oracle labels as well \cite{tramer2020fundamental}, as shown in Fig.\ref{fig:lp_pert}(g), (i) and (j). Naively scaling existing Adversarial Training algorithms to large perturbation bounds would enforce consistent labels on images that flip the Oracle prediction as well, leading to a conflict in the training objective as shown in Fig.\ref{fig:issue with AT}. This results in a large drop in clean accuracy, as shown in Table-\ref{table:cifar10_16_main}. This has triggered interest towards developing perceptually aligned threat models, and defenses that are robust under these settings \cite{laidlaw2020perceptual}. However, finding a perceptually aligned metric is as challenging as building a network that can replicate oracle predictions \cite{tramer2020fundamental}. Thus, it is crucial to investigate adversarial robustness using the well-defined $\ell_p$ norm metric under larger perturbation bounds.

In this work, we aim to improve robustness at larger epsilon bounds, such as an $\ell_\infty$ norm bound of $16/255$ on the CIFAR-10 and CIFAR-100 datasets \cite{krizhevsky2009learning}. 

We define this as a moderate-magnitude bound, and discuss the ideal goals for achieving robustness under this threat model in Sec.\ref{sec:obj_pd}. We further propose a novel defense Oracle-Aligned Adversarial Training (OA-AT), which attempts to align the predictions of the network with that of an Oracle, rather than enforcing all samples within the constraint set to have the same label as the original image.

%\vspace{0.2cm}

\noindent Our contributions have been summarized below:
\begin{itemize}
\itemsep0em
%\vspace{-0.1cm}

    \item[$\bullet$] We propose Oracle-Aligned Adversarial Training (OA-AT) to improve robustness within the defined moderate-$\varepsilon$ threat model.
    \item[$\bullet$] We demonstrate superior performance when compared to state-of-the-art methods such as AWP \cite{wu2020adversarial}, TRADES \cite{zhang2019theoretically} and PGD-AT \cite{madry-iclr-2018} at $\varepsilon=16/255$ while also performing better at $\varepsilon=8/255$ on CIFAR-10 and SVHN. We also demonstrate improved performance on challenging datasets such as CIFAR-100 and Imagenette (10-class subset of Imagenet with 160$\times$160 images).
    \item[$\bullet$] We achieve improvements over the baselines even at larger model capacities such as WideResNet-34-10, and demonstrate results that outperform existing methods on the RobustBench leaderboard. 
    \item[$\bullet$] We show the relation between contrast level of images and the existence of attacks that can flip the Oracle label within a given perturbation bound, and use this observation for constructing better evaluation metrics at large perturbation bounds. 

\end{itemize}

Our code is available here: \url{https://github.com/val-iisc/OAAT}.

\section{Related Works}
\label{sec:rel_works}

\textbf{Robustness against imperceptible attacks:} 
Following the discovery of adversarial examples by Szegedy et al., \cite{intriguing-iclr-2014}, a myriad of adversarial attack and defense methods have been proposed. Adversarial Training has emerged as the most successful defense strategy against $\ell_p$ norm bound imperceptible attacks. PGD Adversarial Training (PGD-AT) proposed by Madry et al. \cite{madry-iclr-2018} constructs multi-step adversarial attacks by maximizing Cross-Entropy loss within the considered threat model and subsequently minimizes the same for training. 

This was followed by several adversarial training methods \cite{zhang2019theoretically,zhang2020attacks,rice2020overfitting,wu2020adversarial,sriramanan2020gama,pang2020bag} that improved accuracy against such imperceptible threat models further. 

Zhang et al. \cite{zhang2019theoretically} proposed the TRADES defense, which maximizes the Kullback-Leibler (KL) divergence between the softmax outputs of adversarial and clean samples for attack generation, and minimizes the same in addition to the Cross-Entropy loss on clean samples for training.

\textbf{Improving Robustness of base defenses:} 
Wu et al. \cite{wu2020adversarial} proposed an additional step of Adversarial Weight Perturbation (AWP) to maximize the training loss, and further train the perturbed model to minimize the same. This generates a flatter loss surface \cite{stutz2021relating}, thereby improving robust generalization. While this can be integrated with any defense, AWP-TRADES is the state-of-the-art adversarial defense today. 

On similar lines, the use of stochastic weight averaging of model weights \cite{izmailov2018averaging} is also seen to improve the flatness of loss surface, resulting in a boost in adversarial robustness  \cite{gowal2020uncovering,chen2021robustov}. Recent works attempt to use training techniques such as early stopping \cite{rice2020overfitting}, optimal weight decay \cite{pang2020bag}, Cutmix data augmentation \cite{yun2019cutmix,rebuffi2021fixing} and label smoothing \cite{rebuffi2021fixing} to achieve enhanced robust performance on base defenses such as PGD-AT \cite{madry-iclr-2018} and TRADES \cite{zhang2019theoretically}. We utilize some of these methods in our approach (Sec.\ref{sec:exp}), and also present improved baselines by combining AWP-TRADES \cite{wu2020adversarial} with these enhancements. %the strongest defense,

\textbf{Robustness against large perturbation attacks:} Shaeiri et al. \cite{shaeiri2020towards} demonstrate that the standard formulation of adversarial training is not well-suited for achieving robustness at large perturbations, as the loss saturates very early. The authors propose Extended Adversarial Training (ExAT), where a model trained on low-magnitude perturbations ($\varepsilon=8/255$) is fine-tuned with large magnitude perturbations ($\varepsilon=16/255$) for just $5$ training epochs, to achieve improved robustness at large perturbations. The authors also discuss the use of a varying epsilon schedule to improve training convergence. 
Friendly Adversarial Training (FAT) \cite{zhang2020attacks} performs early-stopping of an adversarial attack by thresholding the number of times the model misclassifies the image during attack generation. The threshold is increased over training epochs to increase the strength of the attack over training. Along similar lines, Sitawarin et al. \cite{sitawarin2020improving} propose Adversarial Training with Early Stopping (ATES), which performs early stopping of a PGD attack based on the margin (difference between true and maximum probability class softmax outputs) of the perturbed image being greater than a threshold that is increased over epochs. 
We compare against these methods and improve upon them significantly using our proposed approach (Sec.\ref{sec:prop_meth}).

\section{Preliminaries and Threat Model}
\label{sec:prelims}
\subsection{Notation}
\label{subsec:notation}
We consider an $N$-class image classification problem with access to a labelled training dataset $\mathcal{D}$. The input images are denoted by $x\in\mathcal{X}$ and their corresponding labels are denoted as $y\in\{1,...,N\}$. The function represented by the Deep Neural Network is denoted by $f_\theta$ where $\theta\in\Theta$ denotes the set of network parameters. The $N$-dimensional softmax output of the input image $x$ is denoted as $f_\theta(x)$. 
Adversarial examples are defined as images that are crafted specifically to fool a model into making an incorrect prediction \cite{goodfellow_papernot}. An adversarial image corresponding to a clean image $x$ would be denoted as $\widetilde{x}$. 
The set of all images within an $\ell_p$ norm ball of radius $\varepsilon$ is defined as $\mathcal{S}(x)=\{\hat{x}: ||\hat{x}-x||_p < \varepsilon\}$. 

In this work, we specifically consider robustness to $\ell_\infty$ norm bound adversarial examples. We define the Oracle prediction of a sample $x$ as the label that a human is likely to assign to the image, and denote it as $O(x)$. For a clean image, $O(x)$ would correspond to the true label $y$, while for a perturbed image it could differ from the original label. 
% \vspace{-0.2cm}  

\subsection{Nomenclature of Adversarial Attacks}
\label{sec:classification}

Tramer et al. \cite{tramer2020fundamental} discuss the existence of two types of adversarial examples: Sensitivity-based examples, where the model prediction changes while the Oracle prediction remains the same as the unperturbed image, and Invariance-based examples, where the Oracle prediction changes while the model prediction remains unchanged.
Models trained using standard empirical risk minimization are susceptible to sensitivity-based attacks, while models which are overly robust to large perturbation bounds could be susceptible to invariance-based attacks. 
Since these definitions are model-specific, we define a different nomenclature which only depends on the input image and the threat model considered: 

\begin{itemize}
%\vspace{-0.15cm}
\itemsep 0em
    \item[$\bullet$] Oracle-Invariant set $OI(x)$ is defined as the set of all images within the bound $\mathcal{S}(x)$, that preserve Oracle label. Oracle is invariant to such attacks: 
    \begin{equation}
    OI(x) := \{\hat{x}: O(\hat{x})= O(x), \hat{x}\in \mathcal{S}(x)\}
    \end{equation}
    \item[$\bullet$] Oracle-Sensitive set $OS(x)$ is defined as the set of all images within the bound $\mathcal{S}(x)$, that flip the Oracle label. Oracle is sensitive to such attacks: 
    \begin{equation}
    OS(x) := \{\hat{x}: O(\hat{x})\neq O(x), \hat{x}\in \mathcal{S}(x)\}
    \end{equation}
% \vspace{-0.1cm}  
\end{itemize}

\subsection{Objectives of the Proposed Defense}
\label{sec:obj_pd}
Defenses based on the conventional $\ell_p$ norm threat model attempt to train models which are invariant to all samples within $\mathcal{S}(x)$. This is an ideal requirement for low $\varepsilon$-bound perturbations, where the added noise is imperceptible, and hence all samples within the threat model are Oracle-Invariant. An example of a low $\varepsilon$-bound constraint set is the $\ell_{\infty}$ threat model with $\varepsilon=8/255$ for the CIFAR-$10$ dataset, which produces adversarial examples that are perceptually similar to the corresponding clean images, as shown in Fig.\ref{fig:lp_pert}(b).

As we move to larger $\varepsilon$ bounds, Oracle-labels begin to change, as shown in Fig.\ref{fig:lp_pert}(g, i, j). For a very high perturbation bound such as $32/255$, the changes produced by an attack are clearly perceptible and in many cases flip the Oracle label as well. Hence, robustness at such large bounds is not of practical relevance. The focus of this work is to achieve robustness within a moderate-magnitude $\ell_p$ norm bound, where some perturbations look partially modified (Fig.\ref{fig:lp_pert}(g)), while others look unchanged (Fig.\ref{fig:lp_pert}(c, e)), as is the case with $\varepsilon=16/255$ for CIFAR-10. The existence of attacks that do not significantly change the perception of the image necessitates the requirement of robustness within such bounds, while the existence of partially Oracle-Sensitive samples makes it difficult to use standard adversarial training methods on the same. The ideal goals for training defenses under this moderate-magnitude threat model are described below:
% \vspace{-0.2cm}
\begin{itemize}
\itemsep0em
    \item[$\bullet$] Robustness against samples which belong to $OI(x)$
    \item[$\bullet$] Sensitivity towards samples which belong to $OS(x)$, with model's prediction matching the Oracle label
    \item[$\bullet$] No specification on samples which cannot be assigned an Oracle label.
% \vspace{-0.6cm}
\end{itemize}

\noindent Given the practical difficulty in assigning Oracle labels during training and evaluation, we consider the following subset of these ideal goals in this work: 
% \vspace{-0.2cm}
\begin{itemize}
\itemsep0em
    \item[$\bullet$] Robustness-Accuracy trade-off, measured using accuracy on clean samples and robustness against valid attacks within the threat model %(discussed below)
    \item[$\bullet$] Robustness against all attacks within an imperceptible radius ($\varepsilon=8/255$ for CIFAR-10), measured using strong white-box attacks \cite{croce2020reliable,sriramanan2020gama}
    \item[$\bullet$] Robustness to Oracle-Invariant samples within a larger radius ($\varepsilon=16/255$ for CIFAR-10), measured using gradient-free attacks \cite{andriushchenko2019square,chen2020rays} 
% \vspace{-0.2cm}
\end{itemize}

% \vspace{-0.1cm} 

\begin{figure*}[t]
\centering
\includegraphics[width=\linewidth]{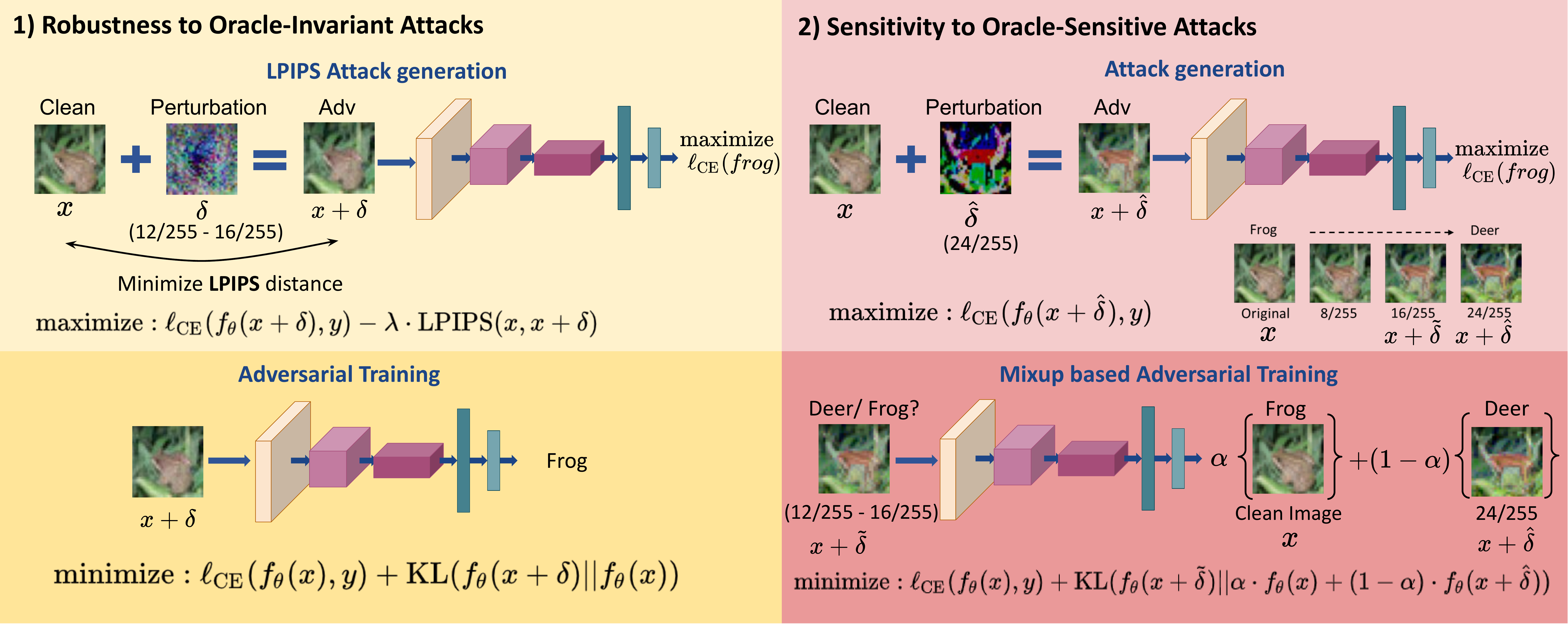}
\caption{\textbf{Oracle-Aligned Adversarial Training:} The proposed defense OA-AT involves alternate training on Oracle-Invariant and Oracle-Sensitive samples. 1) Oracle-Invariant samples are generated by minimizing the LPIPS distance between the clean and perturbed images in addition to the maximization of the Classification Loss. 2) Oracle-Sensitive samples are trained using a convex combination of the predictions of the clean image and the perturbed image at a larger perturbation bound as reference in the KL divergence loss.}
\label{fig:oaat}
%\vspace{-0.25cm}
\end{figure*}

\section{Proposed Method}
\label{sec:prop_meth}
In order to achieve the goals discussed in Sec.\ref{sec:obj_pd}, we require to generate Oracle-Sensitive and Oracle-Invariant samples and impose specific training losses on each of them individually. 
Since labeling adversarial samples as Oracle-Invariant or Oracle-Sensitive is expensive and cannot be done while training networks, we propose to use attacks which ensure a given type of perturbation (OI or OS) by construction, and hence do not require explicit annotation. 

\noindent \textbf{Generation of Oracle-Sensitive examples: } Robust models are known to have perceptually aligned gradients \cite{tsipras2018robustness}. Adversarial examples generated using a robust model tend to look like the target (other) class images at large perturbation bounds, as seen in Fig.\ref{fig:lp_pert}(g, i, j). We therefore use large $\varepsilon$-bound white-box adversarial examples generated from the model being trained as Oracle-Sensitive samples, and the model prediction as a proxy to the Oracle prediction.

\noindent \textbf{Generation of Oracle-Invariant examples: } While the strongest Oracle-Invariant examples are generated using the gradient-free attacks Square \cite{andriushchenko2019square} and Ray-S \cite{chen2020rays}, they require a large number of queries ($5000$ to $10000$), which is computationally expensive for use in adversarial training. Furthermore, reducing the number of queries weakens the attack significantly. The most efficient attack that is widely used for adversarial training is the PGD $10$-step attack. However, it cannot be used for the generation of Oracle-Invariant samples as gradient-based attacks generated from adversarially trained models produce Oracle-Sensitive samples. 
We propose to use the Learned Perceptual Image Patch Similarity (LPIPS) measure for the generation of Oracle-Invariant attacks, as it is known to match well with perceptual similarity based on a study involving human annotators \cite{zhang2018unreasonable,laidlaw2020perceptual}. Further, we observe that while the standard AlexNet model used in prior work \cite{laidlaw2020perceptual} fails to distinguish between Oracle-Invariant and Oracle-Sensitive samples, an adversarially trained model is able to distinguish between the two effectively as shown in Fig.\ref{fig:lpips_plot}. We therefore propose to minimize the LPIPS distance between natural and perturbed images, in addition to the maximization of Cross-Entropy loss for attack generation:  $\mathcal{L}_{CE}(x,y) - \lambda \cdot \textrm{LPIPS}(x,\hat{x})$. The ideal setting of $\lambda$ is the minimum value that transforms attacks from Oracle-Sensitive to Oracle-Invariant (OI) for majority of the images. This results in the generation of strong Oracle-Invariant (OI) attacks. We present several Oracle-Invariant examples for visual inspection in Fig.\ref{fig:lpips_attack_app2}.
\begin{algorithm}[tb]
   \caption{Oracle-Aligned Adversarial Training}
   \label{alg:train_algo}
   
\begin{algorithmic}[1]
   \STATE {\bfseries Input:} Deep Neural Network $f_{\theta}$ with parameters $\theta$, Training Data $\{x_i,y_i \}_{i=1}^{M}$, Epochs $T$, Learning Rate $\eta$, Perturbation budget $\varepsilon_{max}$,  Adversarial Perturbation function $A(x,y,\ell,\varepsilon)$ which maximises loss $\ell$
%   \STATE Initialize $noChange = true$.
\FOR{$\textrm{epoch}=1$ {\bfseries to} $T$}
    \STATE $\widetilde{\varepsilon} = \max \{  \varepsilon_{max}/4, \varepsilon_{max} \cdot \textrm{epoch}/T\}$ 
    % \small{\textcolor{gray}{\#Ramp-up budget}}
    \FOR{$i=1$ {\bfseries to} $M$}
        \STATE $\delta_i \sim U(-\min(\widetilde{\varepsilon},\varepsilon_{max}/4),\min(\widetilde{\varepsilon},\varepsilon_{max}/4))$
        \IF{$\widetilde{\varepsilon} < 3/4\cdot \varepsilon_{max}$}
        \STATE $\ell = \ell_{CE}(f_{\theta}(x_i+\delta_i),y_i)$ ~,~  $\widetilde{\delta}_i = A(x_i,y_i,\ell,\widetilde{\varepsilon})$
        \STATE $L_{adv}= \KL \big(  f_\theta(x_i+\widetilde{\delta}_i)\vert\vert f_\theta(x_i)  \big)$
        \ELSIF{$i\mathbin{\%} 2=0$}
            \STATE $\ell = \ell_{CE}(f_{\theta}(x_i+\delta_i),y_i)$ ~,~  $\widehat{\delta_i} = A(x_i,y_i,\ell, \varepsilon_{ref})$ ~,~  $\widetilde{\delta}_i = \Pi_{\infty}(\widehat{\delta_i},\widetilde{\varepsilon} )$
        \STATE $L_{adv}$ ~=~ $\KL$  \big($ f_\theta(x_i+\widetilde{\delta}_i) ~ \vert\vert ~  \alpha \cdot f_\theta(x_i) + (1-\alpha)\cdot f_\theta(x_i+\widehat{\delta_i})$ \big) 
        
        \ELSE 
            \STATE $\delta_i \sim U(-\widetilde{\varepsilon},\widetilde{\varepsilon}~)$
            \STATE $\ell = \ell_{CE}(f_{\theta}(x_i+\delta_i),y_i)  - \LPIPS(x_i,x_i+\delta_i$),  $\widetilde{\delta_i} = A(x_i,y_i,\ell,\widetilde{\varepsilon})$
            \STATE $L_{adv}= \KL \big(  f_\theta(x_i+\widetilde{\delta}_i)~\vert\vert~ f_\theta(x_i)  \big)$
            
        \ENDIF

        \STATE $L = \ell_{CE}(f_{\theta} (x_i) , y_i ) + L_{adv}$

        \STATE $\theta = \theta -  \eta \cdot \nabla_{\theta} L   $

    \ENDFOR

\ENDFOR
\end{algorithmic}
\end{algorithm}

\textbf{Oracle-Aligned Adversarial Training (OA-AT):} The training algorithm for the proposed defense, Oracle-Aligned Adversarial Training (OA-AT) is presented in Algorithm-\ref{alg:train_algo} and illustrated in Fig.\ref{fig:oaat}. We denote the maximum perturbation bound used for attack generation during the training by $\varepsilon_{max}$. We use the AWP-TRADES formulation \cite{zhang2019theoretically,wu2020adversarial} as the base implementation. Similar to Wu et al. \cite{wu2020adversarial}, we use 10 steps of optimization for attack generation and one additional weight perturbation step. We maximize the classification loss on $x_i+2\cdot\widetilde{\delta_i}$ (where $\widetilde{\delta_i}$ is the attack) in the additional weight perturbation step (instead of $x_i+\widetilde{\delta_i}$ \cite{wu2020adversarial}), in order to achieve better smoothness in the loss surface. We start training with attacks constrained within a perturbation bound of $\varepsilon_{max}/4$ upto one-fourth the training epochs (Alg.\ref{alg:train_algo}, L6-L8), and ramp up this value linearly to $\varepsilon_{max}$ at the last epoch alongside a cosine learning rate schedule. The use of a fixed epsilon initially helps in improving the adversarial robustness faster, while the use of an increasing epsilon schedule later results in better training stability \cite{shaeiri2020towards}. We use $5$ attack steps upto $\varepsilon_{max}/4$ to reduce computation and $10$ attack steps later. 

We perform standard adversarial training upto a perturbation bound of $3/4\cdot\varepsilon_{max}$ as the attacks in this range are imperceptible, based on the chosen moderate-magnitude threat model discussed in Sec.\ref{sec:obj_pd}. Beyond this, we start incorporating separate training losses for Oracle-Invariant and Oracle-Sensitive samples in alternate training iterations (Alg.\ref{alg:train_algo}, L9-L15), as shown in Fig.\ref{fig:oaat}. Oracle-Sensitive samples are generated by maximizing the classification loss in a PGD attack formulation. Rather than enforcing the predictions of such attacks to be similar to the original image, we allow the network to be partially sensitive to such attacks by training them to be similar to a convex combination of predictions on the clean image and perturbed samples constrained within a bound of $\varepsilon_{ref}$, which is chosen to be greater than or equal to $\varepsilon_{max}$ (Alg.\ref{alg:train_algo}, L10). This component of the overall training loss is shown below:
\begin{equation}
\label{eq:OS}
KL\big( f_\theta(x_i+\widetilde{\delta}_i)~\vert\vert~ \alpha~f_\theta(x_i) + (1-\alpha)~f_\theta(x_i+\widehat{\delta_i})   \big)
\end{equation}

%\vspace{-0.5cm}
Here $\widetilde{\delta}_i$ is the perturbation at the varying epsilon value $\widetilde{\varepsilon}$, and $\widehat{\delta_i}$ is the perturbation at $\varepsilon_{ref}$. This loss formulation results in better robustness-accuracy trade-off as shown in E1 versus E3 of Table-\ref{table:ablations_all_comb}. In the alternate iteration, we use the LPIPS metric to efficiently generate strong Oracle-Invariant attacks during training (Alg.\ref{alg:train_algo}, L14). We perform exponential weight-averaging of the network being trained and use this for computing the LPIPS metric for improved and stable results (E1 versus E2 and F1 versus F2 in Table-\ref{table:ablations_all_comb}). We therefore do not need additional training or computation time for training this model. We increase $\alpha$ and $\lambda$ over training, as the nature of attacks changes with varying $\widetilde{\varepsilon}$. The use of both Oracle-Invariant (OI) and Oracle-Sensitive (OS) samples ensures robustness to Oracle-Invariant samples while allowing sensitivity to partially Oracle-Sensitive samples.

\section{Analysing Oracle Alignment of Adversarial Attacks}
We first consider the problem of generating Oracle-Invariant and Oracle-Sensitive attacks in a simplified, yet natural setting to enable more fine-grained theoretical analysis. We consider a binary classification task as introduced by Tsipras et al. \cite{tsipras2018robustness}, consisting of data samples $(x,y)$, with $y\in\{+1,-1\}$, $x\in\mathbb{R}^{d+1}$. Further, 
%\vspace{-0.1cm}
$$x_1 = 
\begin{cases}
y,&\text{w.p. } p\\
-y,&\text{w.p. } 1-p\\
\end{cases}~,~ x_i \sim \mathcal{N}(\alpha y,1)~ \forall i \in \{2,\dots,d+1\} $$
%\vspace{-0.1cm}
In this setting, $x_1$ can be viewed as a feature that is strongly correlated with the Oracle Label $y$ when the Bernoulli parameter $p$ is sufficiently large (for eg: $p \approx 0.90$), and thus corresponds to an Oracle Sensitive feature.  On the other hand, $x_2,\dots,x_{d+1}$ are spurious features that are positively correlated (in a weak manner) to the Oracle label $y$, and are thus Oracle Invariant features. Building upon theoretical analysis presented by Tsipras et al. \cite{tsipras2018robustness}, we make a series of observations, whose details we expound in Section-\ref{theorem_suppl}:

\noindent
\textbf{Observation 1.} Adversarial perturbations of a standard, non-robust classifier utilize spurious features, resulting in Oracle Invariant Samples that are weakly anti-correlated with the Oracle label $y$.

\noindent
\textbf{Observation 2.} Adversarial perturbations of a robust model result in Oracle Sensitive Samples, utilizing features strongly correlated with the Oracle label $y$.

%\vspace{2pt}
\noindent

\begin{figure*}[t]
\centering
\resizebox{0.9\linewidth}{!}{
\begin{subfigure}{0.4\linewidth}
        \centering
        \includegraphics[width=0.9\linewidth]{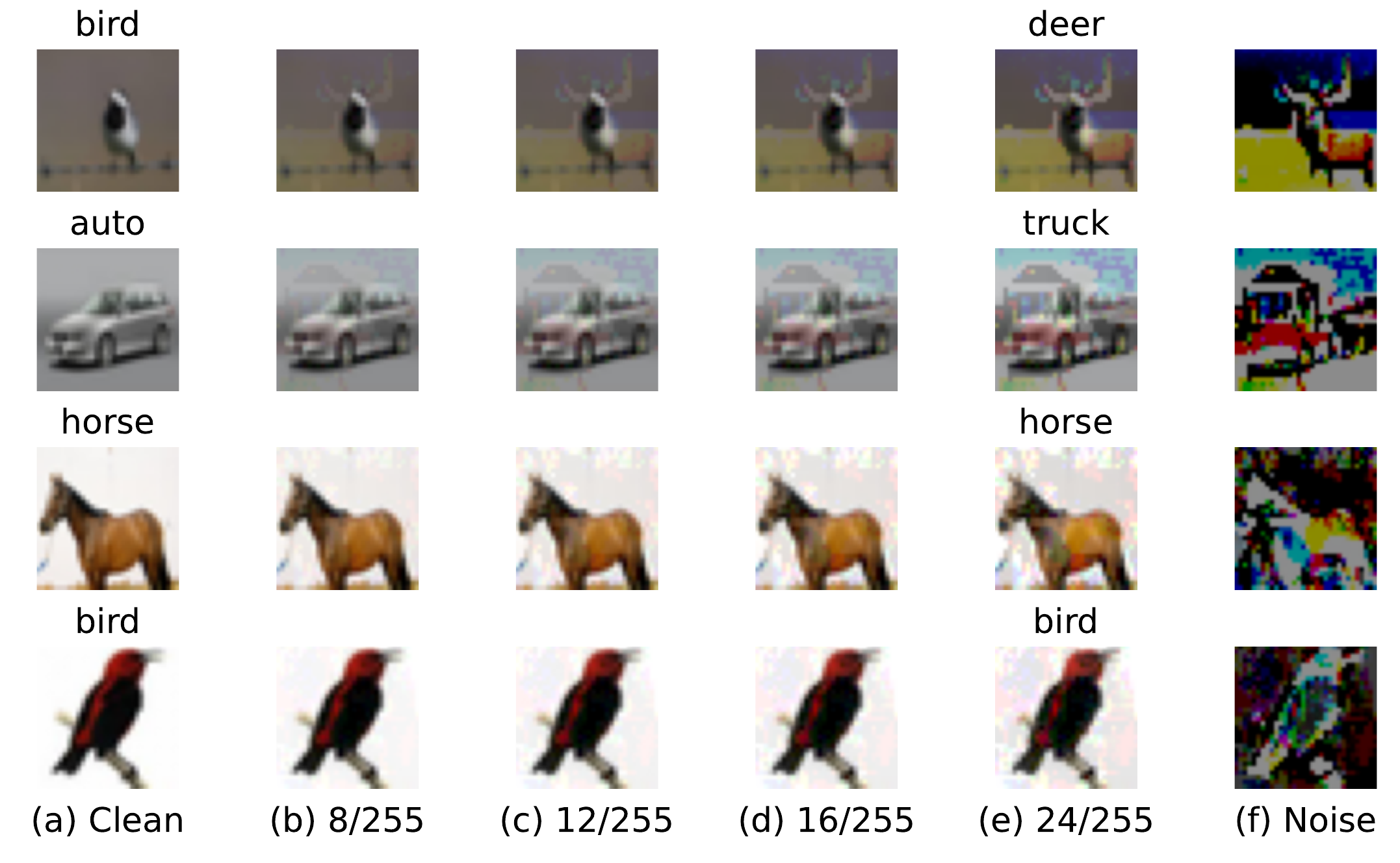}
        \caption{\textbf{CIFAR-10}}
        %\vspace{-0.2cm}
        \label{fig:cifar10_contrast_main}
\end{subfigure}
\hfill
\begin{subfigure}{0.4\linewidth}
        \centering
        \includegraphics[width=0.9\linewidth]{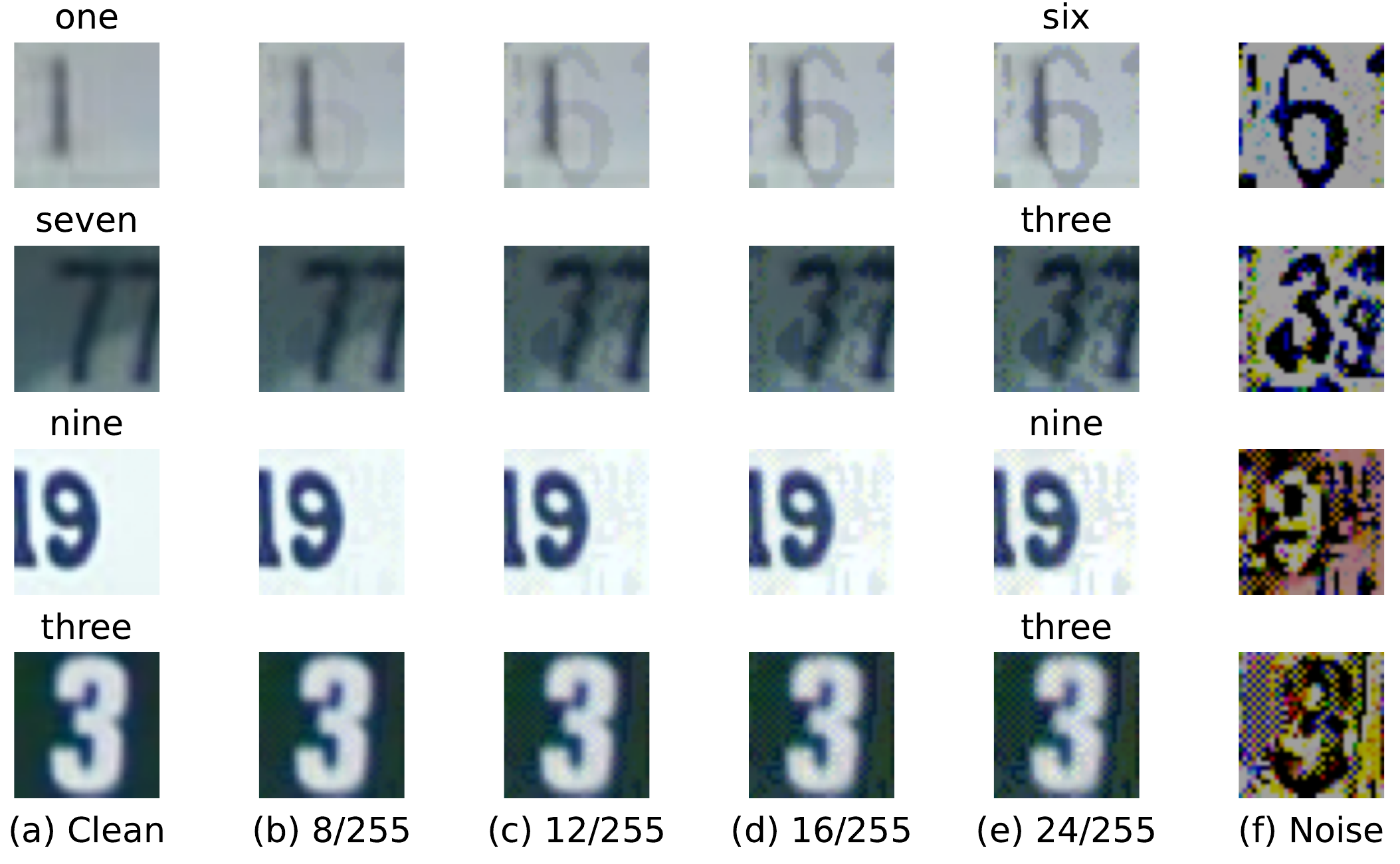}
        \caption{\textbf{SVHN}}
        %\vspace{-0.2cm}
        \label{fig:svhn_contrast_main}
\end{subfigure}}
\caption{\textbf{Relation between the contrast level of an image and the Oracle-Sensitivity of adversarial examples} within a given perturbation bound. First and second rows show low contrast images, and third and fourth rows show high contrast images. Column (a) shows the original clean image and columns (b-e) show adversarial examples at different perturbation bounds generated at the largest bound in (e) and projected to the other bounds in (b, c, d). The adversarial perturbation is shown in column (f). Adversarial examples in columns (d) and (e) are Oracle-Invariant for the high contrast images, and Oracle-Sensitive for the low contrast images.}
\label{fig:contrast_main}
%\vspace{-0.6cm}
\end{figure*}

\section{Role of Image Contrast in Robust Evaluation}
\label{sec:contrast}

As shown in Fig.\ref{fig:lp_pert}, perturbations constrained within a low-magnitude bound  (Fig.\ref{fig:lp_pert}(b)) do not change the perceptual appearance of an image, whereas perturbations constrained within very large bounds such as $\varepsilon=32/255$ (Fig.\ref{fig:lp_pert}(j)) flip the Oracle prediction. As noted by Balaji et al.  \cite{balaji2019instance}, the perturbation radius at which the Oracle prediction changes varies across images. We hypothesize that the contrast level of an image plays an important role in determining the minimum perturbation magnitude $\varepsilon_{OS}$ that can flip the Oracle prediction of an image to generate an Oracle-Sensitive (OS) sample. We visualize a few High-Contrast and Low-Contrast images of the CIFAR-10 and SVHN datasets in Fig.\ref{fig:contrast_main} (more comprehensive visualisations are made available in Fig.\ref{fig:CIFAR10_LC}, \ref{fig:CIFAR10_HC}, \ref{fig:CIFAR100_LC}, \ref{fig:CIFAR100_HC}, \ref{fig:svhn_LC} and \ref{fig:svh_HC}. We observe that High-contrast (HC) images are Oracle-Invariant even at large perturbation bounds, while Low-Contrast (LC) images are Oracle-Sensitive at lower perturbation bounds as well. Based on this, we present robust evaluations at large epsilon bounds on images of varying contrast levels in Fig.\ref{fig:subset_eval}.

\section{Experiments and Results}
\label{sec:exp}
\begin{table*}[t]
\caption{\textbf{Comparison with existing methods:} Performance ($\%$) of the proposed defense OA-AT when compared to baselines against the attacks, GAMA-PGD100 \cite{sriramanan2020gama}, AutoAttack (AA) \cite{croce2020reliable} and an ensemble of Square \cite{andriushchenko2019square} and Ray-S \cite{chen2020rays} attacks (SQ+RS), with different $\varepsilon$ bounds. Sorted by AutoAttack (AA) accuracy at $\varepsilon = 8/255$ for CIFAR-10, CIFAR-100 and Imagenette, and $4/255$ for SVHN. }
\label{table:all_main}
\begin{subtable}{0.46\linewidth}
\caption{\textbf{CIFAR-10, SVHN}}
\setlength\tabcolsep{1pt}
\resizebox{1.0\linewidth}{!}{
\label{table:cifar10_main}
\begin{tabular}{l|cccc|cc}
\toprule
                & \multicolumn{4}{c}{\textbf{Metrics of interest}}                                                                                                                                                                           & \multicolumn{2}{|c}{\textbf{Others}}                                                                                               \\
                \cline{2-7}
\small{\textbf{Method}} & \small{\textbf{~Clean~}} & \textbf{\begin{tabular}[c]{@{}c@{}}\small{~GAMA~}\\ \small{8/255}\end{tabular}} & \textbf{\begin{tabular}[c]{@{}c@{}}\small{~AA~}\\ \small{8/255}\end{tabular}} & \textbf{\begin{tabular}[c]{@{}c@{}}\small{~SQ+RS~}\\ \small{16/255}\end{tabular}} & \textbf{\begin{tabular}[c]{@{}c@{}}\small{~GAMA~}\\ \small{16/255}\end{tabular}} & \textbf{\begin{tabular}[c]{@{}c@{}}\small{~AA~}\\ \small{16/255}\end{tabular}} \\
\midrule
\multicolumn{7}{c}{\textbf{CIFAR-10 (ResNet-18), 110 epochs}}                                                                                                                                                                                                                                                                                                                    \\
\midrule
FAT             & \textbf{84.36}          & 48.41                                                           & 48.14                                                         & 23.22                                                                   & 15.18                                                            & 14.22                                                          \\
PGD-AT          & 79.38          & 49.28                                                           & 48.68                                                         & 25.43                                                                   & 18.18                                                            & 17.00                                                          \\
AWP             & 80.32          & 49.06                                                           & 48.89                                                         & 25.99                                                                   & 19.17                                                            & 18.77                                                          \\
ATES            & 80.95          & 49.57                                                           & 49.12                                                         & 26.43                                                                   & 18.36                                                            & 16.30                                                          \\
TRADES     & 80.53          & 49.63                                                           & 49.42                                                         & 26.20                                                                   & 19.27                                                            & 18.23                                                          \\
ExAT + PGD      & 80.68          & 50.06                                                           & 49.52                                                         & 25.13                                                                   & 17.81                                                            & 19.53                                                          \\
ExAT + AWP   & 80.18          & 49.87                                                           & 49.69                                                         & 27.04                                                                   & 20.04                                                            & 16.67                                                          \\
AWP          & 80.47          & 50.06                                                           & 49.87                                                         & 27.20                                                                   & 19.66                                                            & 19.23                                                          \\
Ours            & 80.24          & \textbf{51.40}                                                           & \textbf{50.88}                                                         & \textbf{29.56}                                                                   & \textbf{22.73}                                                            & \textbf{22.05}                                                          \\
\midrule
\multicolumn{7}{c}{\textbf{CIFAR-10 (ResNet-34), 110 epochs}}                                                                                                                                                                                                                                                                                                                    \\
\midrule
AWP             & 83.89          & 52.64                                                           & 52.44                                                         & 27.69                                                                   & 20.23                                                            & 19.69                                                          \\
OA-AT (Ours)    & \textbf{84.07}          & \textbf{53.54}                                                           & \textbf{53.22}                                                         & \textbf{30.76}                                                                   & \textbf{22.67}                                                            & \textbf{22.00}                                                          \\
\midrule
\multicolumn{7}{c}{\textbf{CIFAR-10 (WRN-34-10), 200 epochs}}                   \\
\midrule
AWP             & 85.36         & 56.34                                                           & 56.17                                                         & 30.87                                                                   & 23.74                                                            & 23.11                                                          \\

OA-AT (Ours)~    & \textbf{85.32}          & \textbf{58.48}                                                           & \textbf{58.04}                                                         & \textbf{35.31}                                                                   & \textbf{26.93}                                                            & \textbf{26.57}            \\

\midrule
\multicolumn{7}{c}{\textbf{SVHN (PreActResNet-18), 110 epochs}}         \\
\midrule
\small{\textbf{Method}} & \small{\textbf{Clean}} & \textbf{\begin{tabular}[c]{@{}c@{}}\small{GAMA}\\ \small{4/255}\end{tabular}} & \textbf{\begin{tabular}[c]{@{}c@{}}\small{AA}\\ \small{4/255}\end{tabular}} & \textbf{\begin{tabular}[c]{@{}c@{}}\small{SQ+RS}\\ \small{12/255}\end{tabular}} & \textbf{\begin{tabular}[c]{@{}c@{}}\small{GAMA}\\ \small{12/255}\end{tabular}} & \textbf{\begin{tabular}[c]{@{}c@{}}\small{AA}\\ \small{12/255}\end{tabular}} \\
\midrule
AWP             & 91.91          & 75.92                                                           & 75.72                                                         & 35.49                                                                   & 30.70                                                            & 30.31                                                          \\
OA-AT (Ours)    & \textbf{94.61}          & \textbf{78.37}                                                           & \textbf{77.96}                                                         & \textbf{39.24}                                                                   & \textbf{34.25}                                                            & \textbf{33.63}               \\

\bottomrule
\end{tabular}}
\end{subtable}
\hfill
\begin{subtable}{0.5\linewidth}
\caption{\textbf{CIFAR-100, ImageNette}}
\setlength\tabcolsep{1pt}
\resizebox{1.0\linewidth}{!}{
\label{table:cifar100_svhn_main}
\begin{tabular}{l|cccc|cc}
\toprule
                & \multicolumn{4}{c}{\textbf{Metrics of interest}}                                                                                                                                                                           & \multicolumn{2}{|c}{\textbf{Others}}                                                                                               \\
                \cline{2-7}
\small{\textbf{Method}} & \small{\textbf{~Clean~}} & \textbf{\begin{tabular}[c]{@{}c@{}}\small{~GAMA~}\\ \small{8/255}\end{tabular}} & \textbf{\begin{tabular}[c]{@{}c@{}}\small{~AA~}\\ \small{8/255}\end{tabular}} & \textbf{\begin{tabular}[c]{@{}c@{}}\small{~SQ+RS~}\\ \small{16/255}\end{tabular}} & \textbf{\begin{tabular}[c]{@{}c@{}}\small{~GAMA~}\\ \small{16/255}\end{tabular}} & \textbf{\begin{tabular}[c]{@{}c@{}}\small{~AA~}\\ \small{16/255}\end{tabular}} \\
\midrule
\multicolumn{7}{c}{\textbf{CIFAR-100 (ResNet-18), 110 epochs}} \\
\midrule
AWP             & 58.81          & 25.51                                                           & 25.30                                                         & 11.39                                                                   & 8.68                                                             & 8.29                                                           \\
AWP+            & 59.88          & 25.81                                                           & 25.52                                                         & 11.85                                                                   & 8.72                                                             & 8.28                                                           \\
OA-AT (no LS)   & 60.27          & 26.41                                                           & 26.00                                                         & 13.48                                                                   & \textbf{10.47}                                                            & \textbf{9.95}                                                           \\
OA-AT (Ours)    & \textbf{61.70}          & \textbf{27.09}                                                           & \textbf{26.77}                                                         & \textbf{13.87}                                                                   & 10.40                                                            & 9.91                                                           \\
\midrule
\multicolumn{7}{c}{\textbf{CIFAR-100 (PreActResNet-18), 200 epochs}}  \\
\midrule
AWP             & 58.85          & 25.58                                                           & 25.18                                                         & 11.29                                                                   & 8.63                                                             & 8.19                                                           \\
AWP+            & \textbf{62.11}          & 26.21                                                           & 25.74                                                         & 12.23                                                                   & 9.21                                                             & 8.55                                                           \\
OA-AT (Ours)    & 62.02          & \textbf{27.45}                                                           & \textbf{27.14}                                                         & \textbf{14.52}                                                                   & \textbf{10.64}                                                            & \textbf{10.10}                                                          \\
\midrule
\multicolumn{7}{c}{\textbf{CIFAR-100 (WRN-34-10), 110 epochs}}   \\
\midrule
AWP             & 62.41          & 29.70                                                           & 29.54                                                         & 14.25                                                                   & 11.06                                                            & 10.63                                                          \\
AWP+            & 62.73          & 29.92                                                           & 29.59                                                         & 14.96                                                                   & 11.55                                                            & 11.04                                                          \\
OA-AT (no LS)~ &	65.22&	30.75&	\textbf{30.35}&	16.77&	12.65 &	11.95 \\
OA-AT (Ours)    & \textbf{65.73}          & \textbf{30.90}                                                           & \textbf{30.35}                                                         & \textbf{17.15}                                                                   & \textbf{13.21}                                                            & \textbf{12.01}                                                          \\

\midrule
\multicolumn{7}{c}{\textbf{Imagenette (ResNet-18), 110 epochs}}         \\
\midrule
\small{\textbf{Method}} & \small{\textbf{Clean}} & \textbf{\begin{tabular}[c]{@{}c@{}}\small{GAMA}\\ \small{8/255}\end{tabular}} & \textbf{\begin{tabular}[c]{@{}c@{}}\small{AA}\\ \small{8/255}\end{tabular}} & \textbf{\begin{tabular}[c]{@{}c@{}}\small{SQ+RS}\\ \small{16/255}\end{tabular}} & \textbf{\begin{tabular}[c]{@{}c@{}}\small{GAMA}\\ \small{16/255}\end{tabular}} & \textbf{\begin{tabular}[c]{@{}c@{}}\small{AA}\\ \small{16/255}\end{tabular}} \\
\midrule

AWP                                                                                    & 82.73          & 57.52                                                           & 57.40   &42.52                                                         & 29.14                                                               &28.86     \\
OA-AT (Ours) & \textbf{82.98}          & \textbf{59.51}                                                           & \textbf{59.31}                                                             & \textbf{48.01}                                                                & \textbf{48.66}
                                                & \textbf{31.78}\\
\bottomrule

\end{tabular}}
\end{subtable}
%\vspace{-0.2cm}
\end{table*}
%\vspace{-0.2cm}
In this section, we present detailed robust evaluations of the proposed approach along with various existing defenses on the CIFAR-10 \cite{krizhevsky2009learning}, CIFAR-100 \cite{krizhevsky2009learning}, SVHN \cite{svhn} and Imagenette \cite{imagenette} datasets. We report adversarial robustness against the strongest known attacks, AutoAttack (AA) \cite{croce2020reliable} and GAMA PGD-100 (GAMA) \cite{sriramanan2020gama} for $\varepsilon=8/255$ in order to obtain the worst-case robust accuracy. For larger bounds such as $12/255$ and $16/255$, we primarily aim for robustness against an ensemble of the Square \cite{andriushchenko2019square} and Ray-S \cite{chen2020rays} attacks, as they generate strong Oracle-Invariant examples. On the SVHN dataset, we find that the perturbation bound for imperceptible attacks is $\varepsilon=4/255$, and consider robustness within $12/255$ (Fig.\ref{fig:svhn_LC} and \ref{fig:svh_HC}).

For each baseline on CIFAR-10, we find the best set of hyperparameters to achieve clean accuracy of around $80\%$ to ensure a fair comparison across all methods. We further perform baseline training across various $\varepsilon$ values and report the best results in  Table-\ref{table:cifar10_main}. We note that existing defenses do not perform well when trained using large $\varepsilon$ bounds such as $16/255$ as shown in Table-\ref{table:cifar10_16_main} (more detailed results available in Table-\ref{table:cifar10_full} and \ref{table:cifar100_full}). On other datasets, we present comparative analysis primarily with AWP \cite{wu2020adversarial}, the leading defense amongst prior methods on the RobustBench Leaderboard \cite{croce2021robustbench} in the setting without additional or synthetic training data, which we consider in this work. We further compare the proposed approach with the AWP baseline using various model architectures (ResNet-18, ResNet-34 \cite{he2016deep}, WideResNet-34-10 \cite{zagoruyko2016wide} and PreActResNet-18 \cite{preact_resnet}). 

Contrary to prior works \cite{rice2020overfitting,rebuffi2021fixing}, we obtain additional gains with the use of the augmentation technique, AutoAugment \cite{cubuk2018autoaugment}. We also use Model Weight Averaging (WA) \cite{izmailov2018averaging,gowal2020uncovering,chen2021robustov} to obtain better generalization performance, especially at larger model capacities. To ensure a fair comparison, we use these methods to obtain improved baselines as well, and report this as AWP+ in Table-\ref{table:all_main} if any improvement is observed (more comprehensive results in Sec.\ref{AWP_plus_suppl}). As observed by Rebuffi et al. \cite{rebuffi2021fixing}, we find that label-smoothing and the use of warmup in the learning rate scheduler helps achieve an additional boost in robustness. However, we report our results without including this as well (no LS) to highlight the gains of the proposed method individually.
\begin{table}[t]
\caption{\textbf{Comparison with RobustBench Leaderboard  \cite{croce2021robustbench} Results:} Performance ($\%$) of the proposed method (OA-AT) when compared to AWP \cite{wu2020adversarial}, which is the state-of-the-art amongst methods that do not use additional training data/ synthetic data on the RobustBench Leaderboard.}
%\vspace{-0.05cm}
\centering
\setlength\tabcolsep{3pt}
\resizebox{0.85\linewidth}{!}{
\label{table:lp_norm}
\begin{tabular}{l|c|c|c|c|c|c|c|c}

\toprule                                                                                                                                                                             
\cmidrule(l){2-9}
\textbf{\small{Method}}                                               & \textbf{\begin{tabular}[c]{@{}c@{}}Clean \\ Acc\end{tabular}}       & \textbf{\begin{tabular}[c]{@{}c@{}}\boldmath$\ell_\infty$ (AA) \\ 8/255\end{tabular}} & \textbf{\begin{tabular}[c]{@{}c@{}}\boldmath$\ell_\infty$ (OI)\\ 16/255\end{tabular}} & \textbf{\begin{tabular}[c]{@{}c@{}}\boldmath$\ell_2$ (AA) \\ \boldmath$\varepsilon=0.5$\end{tabular}} & 
\textbf{\begin{tabular}[c]{@{}c@{}}\boldmath$\ell_2$ (AA) \\ \boldmath$\varepsilon=1$\end{tabular}} & 
\textbf{\begin{tabular}[c]{@{}c@{}}\boldmath$\ell_1$ (AA) \\ \boldmath$\varepsilon=5$\end{tabular}} & \textbf{\begin{tabular}[c]{@{}c@{}}\boldmath$\ell_0$ \small{(PGD$_0$)}\\ \boldmath$\varepsilon=7$\end{tabular}} & \textbf{\begin{tabular}[c]{@{}c@{}}Comm\\Corr\end{tabular}} \\
\midrule
                                                                                                                                                                                                       
\multicolumn{9}{c}{\textbf{CIFAR-10 (WRN-34-10)}}                   \\
\midrule                                                                                                                       
AWP                    & \textbf{85.36}                                  & 56.17                                        & 30.87                                                  & 60.68                                    & 28.86                                  & 37.29                                                          & 39.09                                      & 75.83                                           \\
Ours                                & 85.32                                  & \textbf{58.04}                                        & \textbf{35.31}                                                  & \textbf{64.08}                                    & \textbf{34.54}                                  & \textbf{45.72}                                                          & \textbf{44.40}                                       & \textbf{76.78}                                           \\

\midrule
                                                                                                                                                                                                       
\multicolumn{9}{c}{\textbf{CIFAR-100 (WRN-34-10)}}                   \\
\midrule     

AWP                                & 62.73                                  & 29.59                                        & 14.96                                                  & 36.62                                    & 17.05                                  & 21.88                                                          & 17.40                                       & 50.73   \\
Ours                                & \textbf{65.73}                                  & \textbf{30.35}                                        & \textbf{17.15}                                                  & \textbf{37.21}                                    & \textbf{17.41}                                  & \textbf{25.75}                                                          & \textbf{29.20}                                       & \textbf{54.88}                                           \\
\bottomrule
\end{tabular}}

%\vspace{-0.5cm}
\end{table}

\begin{figure*}[t]
\centering
\includegraphics[width=\linewidth]{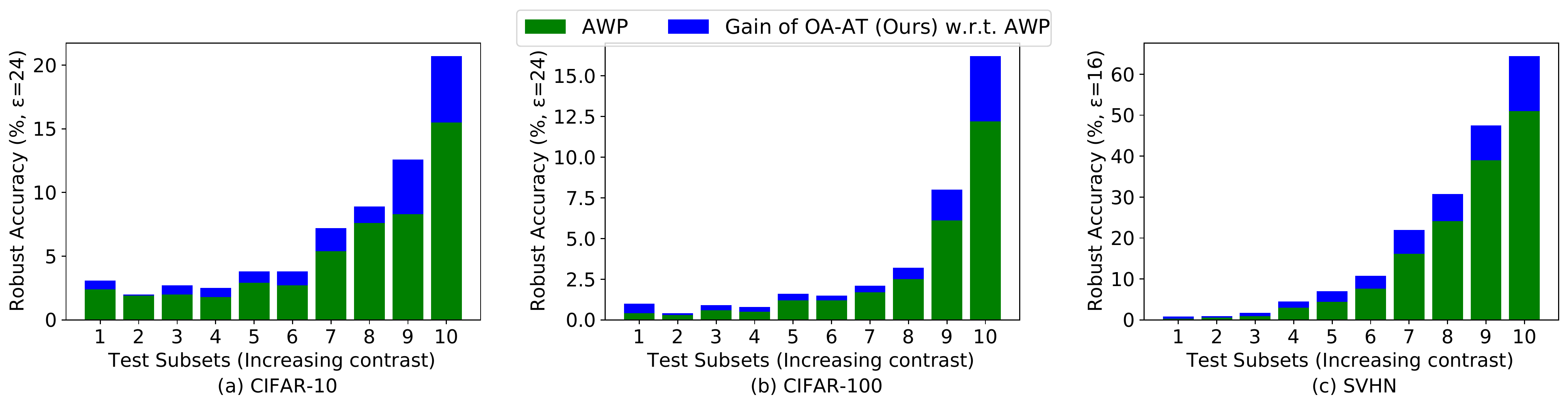}
%\vspace{-0.7cm}
\caption{\textbf{Evaluation across test subsets of increasing contrast levels:} Here we plot the gain in robust accuracy of the proposed defense OA-AT over AWP \cite{wu2020adversarial}. The proposed defense achieves higher gains as contrast increases, verifying that the proposed approach is more robust to the Oracle-Invariant white-box attacks on High-Contrast images.}
\label{fig:subset_eval}
%\vspace{-0.45cm}
\end{figure*}

From Table-\ref{table:all_main}, we observe that the proposed defense achieves significant and consistent gains across all metrics specified in Sec.\ref{sec:obj_pd}. The proposed approach outperforms existing defenses by a significant margin on all four datasets, over different network architectures. Although we train the model for achieving robustness at larger $\varepsilon$ bounds, we achieve an improvement in the robustness at the low $\varepsilon$ bound (such as $\varepsilon=8/255$ on CIFAR-10) as well, which is not observed in any existing method (Sec.\ref{detailed_res_supp}). We also report the results on $\ell_2$ Norm adversaries in Table-\ref{table:l2_PRN18_mini}. As shown in Fig.\ref{fig:subset_eval}, the proposed defense achieves higher gains on the high contrast test subsets of different datasets, verifying that the proposed approach has better robustness against Oracle-Invariant attacks, and not against Oracle-Sensitive attacks. 

\textbf{RobustBench Leaderboard Comparisons: } As shown in Table-\ref{table:lp_norm}, using the proposed method, we obtain a significant improvement over state-of-the-art results reported on the RobustBench Leaderboard (AWP) without the use of additional/ synthetic data on both CIFAR-10 and CIFAR-100 datasets. We observe that the proposed approach achieves significant gains against $\ell_\infty$ norm bound attacks at $\varepsilon=8/255$ and $16/255$ that were used for training, as well as other $\ell_p$ norm bound attacks and common corruptions on both datasets. 

\begin{figure*}[t]
%\vspace{-0.3cm}
% \begin{minipage}{0.48\linewidth}
\centering
        \includegraphics[width=\linewidth]{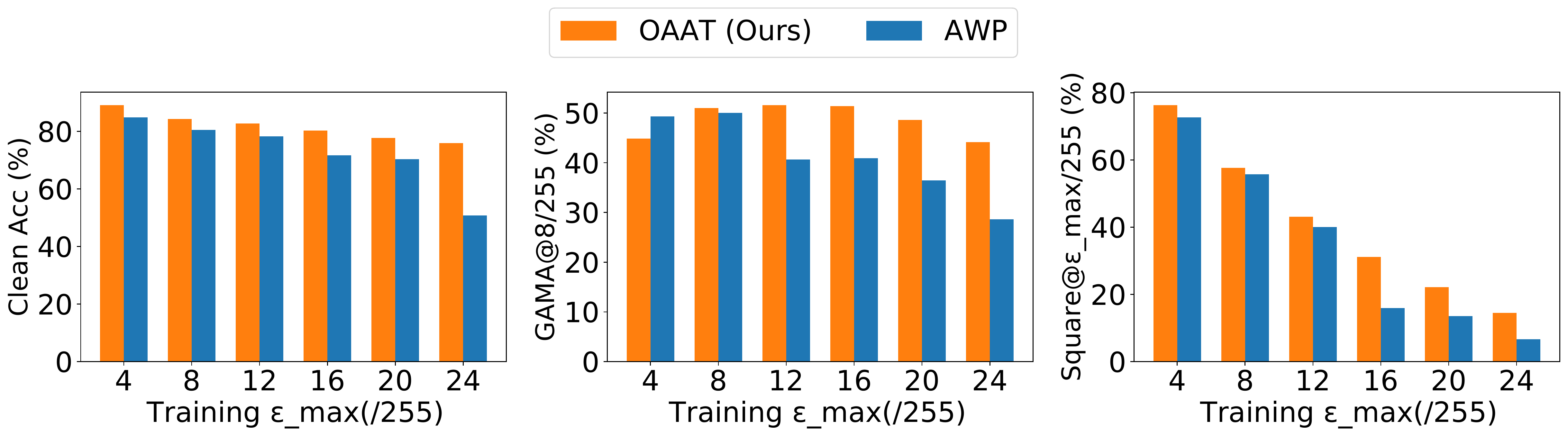}
        %\vspace{-0.4cm}
        \caption{\textbf{Results across variation in training $\varepsilon_{max}$:} While the proposed approach works best at moderate-$\varepsilon$ bounds such as 16/255 on CIFAR-10, we observe that it outperforms the baseline for various $\varepsilon_{max}$ values $\geq8/255$ as well.}
        \label{fig:acc_maxeps_curve}
        % \vspace{-0.5cm}
        % \end{minipage}
        % \hfill

%\vspace{-0.5cm}
\end{figure*}

The $\varepsilon_{max}$ used for training is a system specification, which is the perturbation bound within which the model has to be robust. Thus, to validate the efficacy of the proposed approach, we train different ResNet-18 models on CIFAR-10 using different specifications of $\varepsilon_{max}$. From Fig.\ref{fig:acc_maxeps_curve}, we observe that for various values of training $\varepsilon_{max}$, the proposed approach consistently outperforms AWP \cite{wu2020adversarial}.
Training time of OA-AT is comparable with that of AWP \cite{wu2020adversarial}. On CIFAR-10, OA-AT takes 7 hours 16 minutes, while AWP takes 7 hours 27 minutes for 110 epochs of training on ResNet-18 using a single V100 GPU. To ensure the absence of gradient masking in the proposed approach, we present further evaluations against diverse attacks and sanity checks in Sec.\ref{sec:grad_mask_checks}.

\textbf{Ablation Study: }
In order to study the impact of different components of the proposed defense, we present a detailed ablative study using ResNet-18 and WideResNet-34-10 models in Table-\ref{table:ablations_all_comb}. We present results on the CIFAR-10 and CIFAR-100 datasets, with E1 and F1 representing the proposed approach. First, we study the efficacy of the LPIPS metric in generating Oracle-Invariant attacks. In experiment E2, we train a model without LPIPS by setting its coefficient to zero. While the resulting model achieves a slight boost in robust accuracy at $\varepsilon=16/255$ due to the use of stronger attacks for training, there is a considerable drop in clean accuracy, and a corresponding drop in robust accuracy at $\varepsilon=8/255$ as well. We observe a similar trend by setting the value of $\alpha$ to $1$ as shown in E3, and by combining E2 and E3 as shown in E4. We note that E4 is similar to standard adversarial training, where the model attempts to learn consistent predictions in the $\varepsilon$ ball around every data sample. While this works well for large $\varepsilon$ attacks ($16/255$), it leads to poor clean accuracy as shown in Table-\ref{table:cifar10_16_main}. 

We further note that the computation of LPIPS distance using an exponential weight averaged model (E1) results in better performance as compared to using the model being trained (E5). As discussed in Sec.\ref{sec:prop_meth}, we maximize loss on $x_i+2\cdot\widetilde{\delta_i}$ (where $\widetilde{\delta_i}$ is the attack) in the additional weight perturbation step. We present results by using the standard $\varepsilon$ limit for the weight perturbation step as well, in E6. This leads to a drop across all metrics, indicating the importance of using large magnitude perturbations in the weight perturbation step for producing a flatter loss surface that leads to better generalization to the test set. Different from the standard TRADES formulation, we maximize Cross-Entropy loss for attack generation in the proposed method. From E7 we note a drop in robust accuracy since the KL divergence based attack is weaker (Gowal et al. \cite{gowal2020uncovering}). We present further ablative analysis in Sec.\ref{ablations_supp}. 
\begin{table*}[t]
%\vspace{-0.4cm}
\caption{\textbf{CIFAR-10, CIFAR-100}: Ablation experiments on ResNet-18 architecture (E1-E7) and WideResNet-34-10 (F1-F2) architecture to highlight the importance of various aspects in the proposed defense OA-AT. Performance ($\%$) against attacks with different $\varepsilon$ bounds is reported.}
\setlength\tabcolsep{2pt}
\resizebox{1.0\linewidth}{!}{
\label{table:ablations_all_comb}
\begin{tabular}{l|c|c|cc||c|c|cc}
\toprule
                                                              & \multicolumn{4}{c||}{\textbf{CIFAR-10}}                                                                                                                                                                                           & \multicolumn{4}{c}{\textbf{CIFAR-100}}                                                                                                                                                                                    \\
\cmidrule(l){2-9}
\textbf{Method}                                               & \textbf{~~Clean~~}       & \textbf{\begin{tabular}[c]{@{}c@{}}~~GAMA~~\\ (8/255)\end{tabular}} & \textbf{\begin{tabular}[c]{@{}c@{}}~~GAMA~~\\ (16/255)\end{tabular}} & \textbf{\begin{tabular}[c]{@{}c@{}}~Square~ \\ (16/255)\end{tabular}} & \textbf{~~Clean~~} & \textbf{\begin{tabular}[c]{@{}c@{}}~~GAMA~~\\ (8/255)\end{tabular}} & \textbf{\begin{tabular}[c]{@{}c@{}}~~GAMA~~\\ (16/255)\end{tabular}} & \textbf{\begin{tabular}[c]{@{}c@{}}~Square~ \\ (16/255)\end{tabular}} \\
\midrule
\textbf{E1}: OA-AT (Ours)                                      & 80.24                & \textbf{51.40}                                                  & 22.73                                                            & 31.16                                                               & 60.27          & \textbf{26.41}                                                  & 10.47                                                            & 14.60                                                               \\
\textbf{E2}: LPIPS weight = 0                                 & 78.47                & 50.60                                                           & 24.05                                                            & 31.37                                                               & 58.47          & 25.94                                                           & 10.91                                                            & 14.66                                                               \\
\textbf{E3}: Alpha = 1                                        & 79.29                & 50.60                                                           & 23.65                                                            & 31.23                                                               & 58.84          & 26.15                                                           & 10.97                                                            & 14.89                                                               \\
\textbf{E4}: Alpha = 1, LPIPS weight = 0                      & 77.16                & 50.49                                                           & \textbf{24.93}                                                   & \textbf{32.01}                                                      & 57.77          & 25.92                                                           & \textbf{11.33}                                                   & \textbf{15.03}                                                      \\
\textbf{E5}: Using Current model (without WA) for LPIPS       & \textbf{80.50}                & 50.75                                                           & 22.90                                                            & 30.76                                                               & 59.54          & 26.23                                                           & 10.50                                                            & 14.86                                                               \\
\textbf{E6}: Without 2*eps perturbations for AWP              & 79.96                & 50.50                                                           & 22.61                                                            & 30.60                                                               & 60.18          & 26.27                                                           & 10.15                                                            & 14.20                                                               \\

\textbf{E7}: Maximizing KL div in the AWP step                & 81.19                & 49.77                                                           & 21.17                                                            & 29.39                                                               & 59.48          & 25.03                                                           & 7.93                                                             & 13.34                                                               \\

\midrule
\textbf{F1}: OA-AT (Ours) & \textbf{85.32}&	\textbf{58.48}&	26.93&	\textbf{36.93}& \textbf{65.73} & \textbf{30.90} & 13.21 & \textbf{18.47}  \\
\textbf{F2}: LPIPS weight = 0  & 83.47&	57.58&	\textbf{27.21}&	36.68& 63.16 & 30.22 & \textbf{13.59} & 18.42  \\
\bottomrule
\end{tabular}}
%\vspace{-0.4cm}
\end{table*} %consistent with the observation

\vspace{-0.1cm}
\section{Conclusions} 
%\vspace{-0.1cm}
In this paper, we investigate in detail robustness at larger perturbation bounds in an $\ell_p$ norm based threat model. We discuss the ideal goals of an adversarial defense at larger perturbation bounds, identify deficiencies of prior works in this setting and further propose a novel defense, Oracle-Aligned Adversarial Training (OA-AT) that aligns model predictions with that of an Oracle during training. The key aspects of the defense include the use of LPIPS metric for generating Oracle-Invariant attacks during training, and the use of a convex combination of clean and adversarial image predictions as targets for Oracle-Sensitive samples. We achieve state-of-the-art robustness at low and moderate perturbation bounds, and a better robustness-accuracy trade-off. We further show the relation between the contrast level of images and the existence of Oracle-Sensitive attacks within a given perturbation bound. We use this for better evaluation, and highlight the role of contrast of images in achieving an improved robustness-accuracy trade-off. We hope that future work would build on this to construct better defenses and to obtain a better understanding on the existence of adversarial examples. 

\vspace{-0.1cm}
\section{Acknowledgements}
This work was supported by a research grant (CRG/2021/005925) from SERB, DST, Govt. of India. Sravanti Addepalli is supported by Google PhD Fellowship and CII-SERB Prime Minister's Fellowship for Doctoral Research. 
% We are thankful for the support.

\bibliographystyle{splncs04}
\bibliography{references}

\clearpage
\thispagestyle{empty}

\begin{center}
\textbf{\Large Supplementary material}
\end{center}
\stepcounter{myequation}
\stepcounter{myalgorithm}
\stepcounter{myfigure}
\stepcounter{mytable}
\stepcounter{mysection}
\makeatletter
\renewcommand{\theequation}{S\arabic{equation}}
\renewcommand{\thefigure}{S\arabic{figure}}
\renewcommand{\thetable}{S\arabic{table}}
\renewcommand{\thesection}{S\arabic{section}}

\section{Oracle-Invariant Attacks}

\label{app_sec:gen_oi_attacks}
\subsection{Square Attack }The strongest Oracle-Invariant examples are generated using the Square attack \cite{andriushchenko2019square}. Images so generated are Oracle-Invariant since the Square Attack is query-based, and does not utilise gradients from the model for attack generation. However this attack uses $5000$ queries, and is thus computationally expensive. Hence it cannot be directly incorporated for adversarial training, although it is one of the strongest attacks for evaluation purposes. We note that the computationally efficiency can be improved by reducing the number of queries; however it also reduces the effectiveness of the attack significantly. The adversarial images generated using the Square attack and their corresponding perturbations are presented in Fig.\ref{fig:square}.

\subsection{RayS Attack} 
\begin{figure*}
\begin{minipage}{0.48\linewidth}
\centering
        \includegraphics[width=\linewidth]{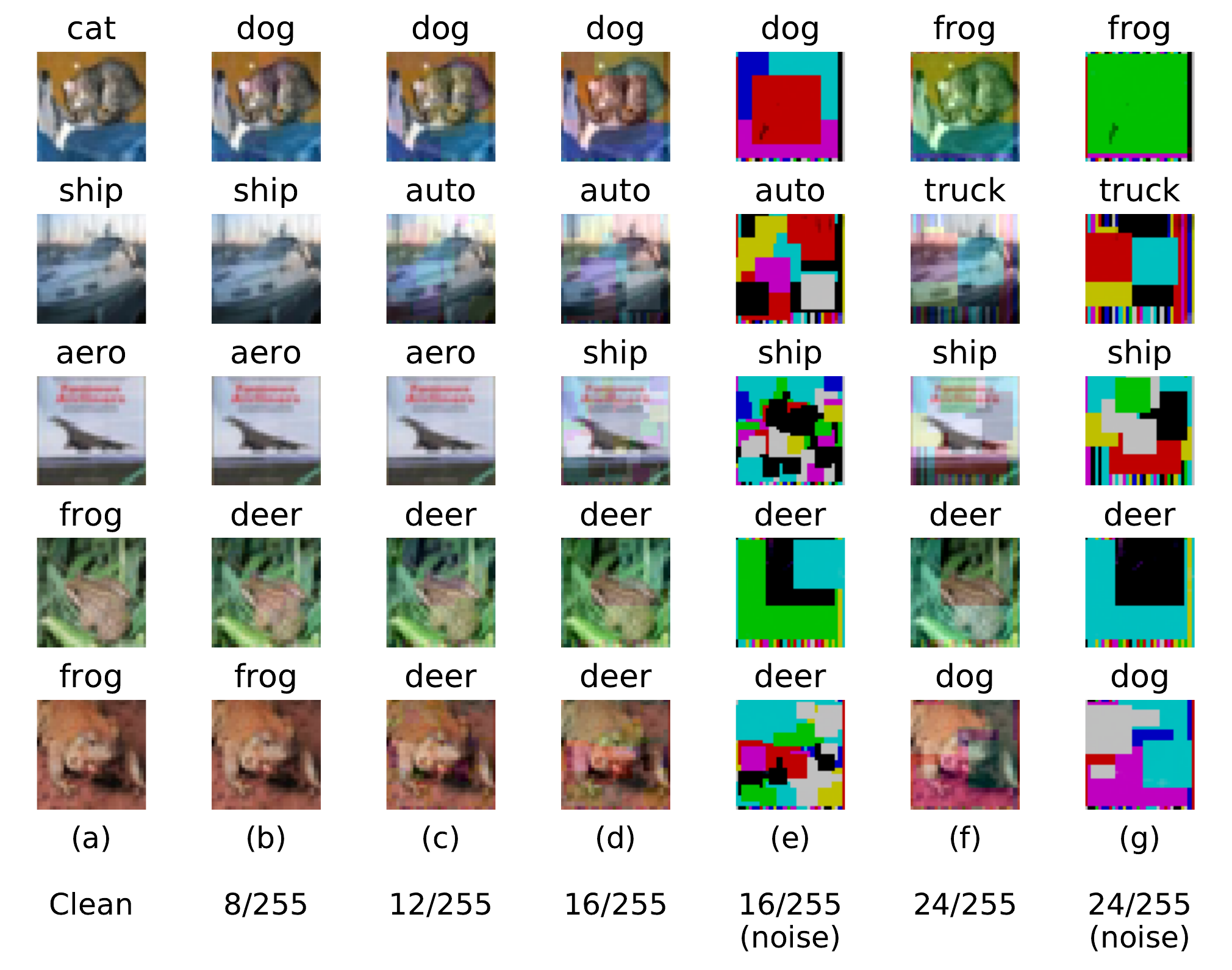}
        % \vspace{-0.2cm}
        \caption{\textbf{Square attack:} Adversarially attacked images (b, c, d, f) and the corresponding perturbations (e, g) for various $\ell_\infty$ bounds generated using the gradient-free random search based attack Square \cite{andriushchenko2019square}. The clean image is shown in (a). Attacks are generated from a model trained using the proposed Oracle-Aligned Adversarial Training (OA-AT) algorithm on CIFAR-10. Prediction of the same model is printed above each image.}
        \label{fig:square}
        % \vspace{-0.2cm}
        \end{minipage}
        \hfill
        \begin{minipage}{0.48\linewidth}
\centering
        \includegraphics[width=\linewidth]{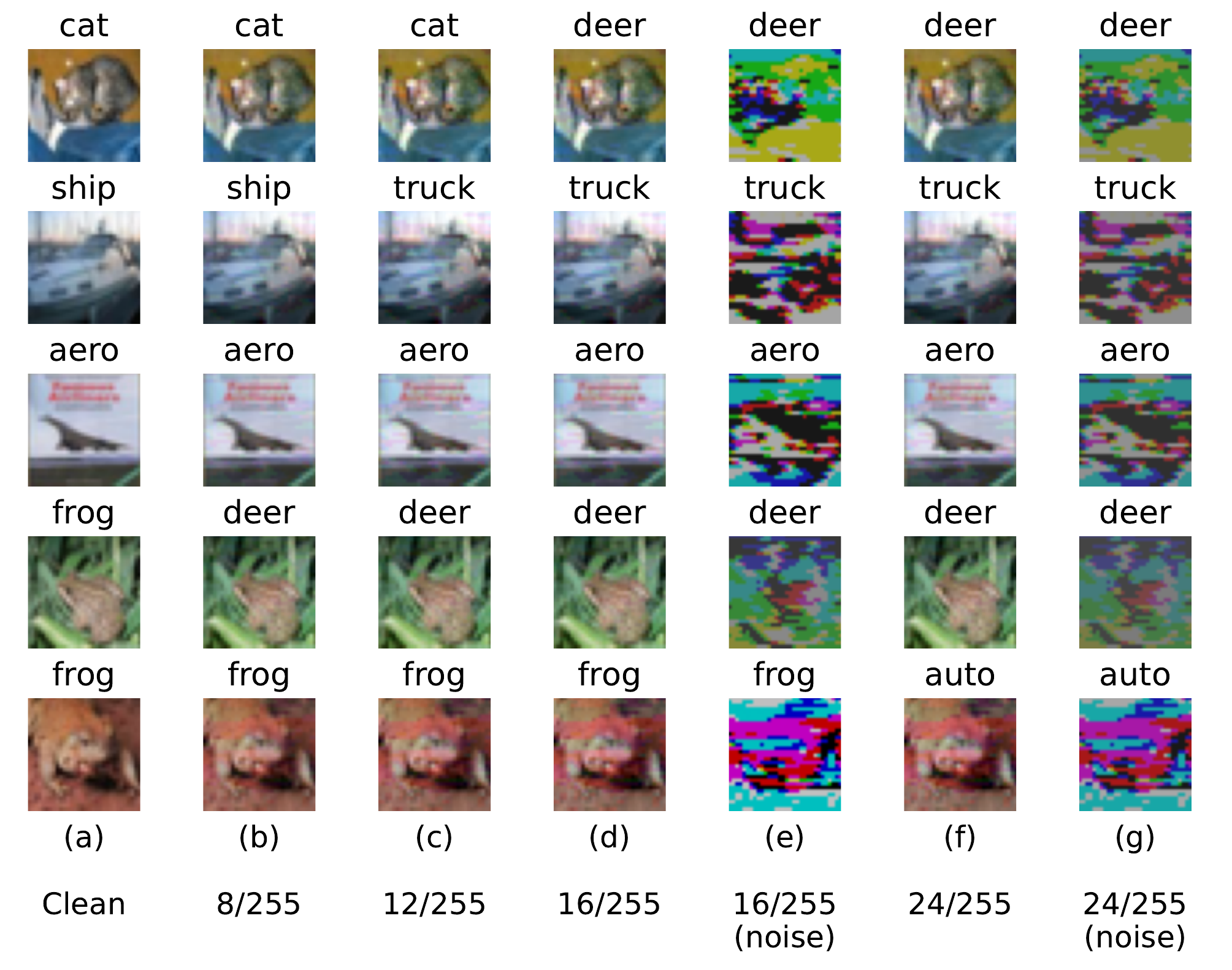}
        % \vspace{-0.2cm}
        \caption{\textbf{RayS attack:} Adversarially attacked images (b, c, d, f) and the corresponding perturbations (e, g) for various $\ell_\infty$ bounds generated using the gradient-free binary search based attack RayS \cite{chen2020rays}. The clean image is shown in (a). Attacks are generated from a model trained using the proposed Oracle-Aligned Adversarial Training (OA-AT) algorithm on CIFAR-10. Prediction of the same model is printed above each image.}
        \label{fig:rayS}
        % \vspace{-0.2cm}
\end{minipage}
        
\end{figure*}

Another technique that is observed to generate strong Oracle-Invariant examples is the black-box RayS attack \cite{chen2020rays}. Similar to the Square attack, the images so generated are also Oracle-Invariant since it is a query-based attack and does not utilise gradients for attack generation. Although the RayS attack requires  $10000$ queries which is highly demanding from a computational viewpoint, it is observed to be weaker than the Square attack. Adversarial images generated using the RayS attack and their corresponding perturbations are presented in Fig.\ref{fig:rayS}.

\begin{figure*}
\begin{minipage}{0.48\linewidth}
\centering
        \includegraphics[width=\linewidth]{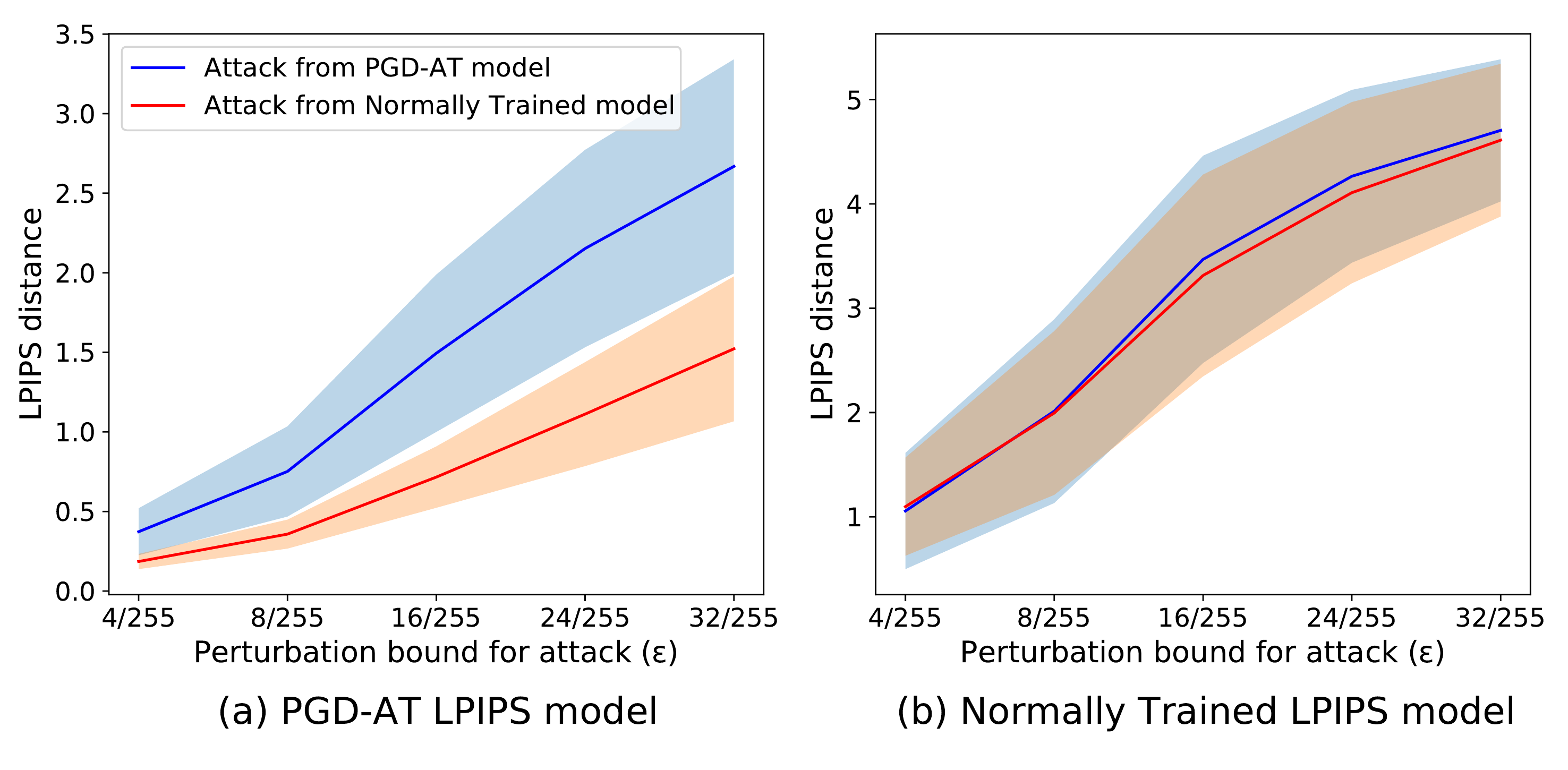}
        % \vspace{-0.5cm}
        \caption{LPIPS distance between clean and adversarially perturbed images. Attacks generated from PGD-AT \cite{madry-iclr-2018,pang2020bag} model (Oracle-Sensitive) and Normally Trained model (Oracle-Invariant) are considered. (a) PGD-AT ResNet-18 model is used for computation of LPIPS distance (b) Normally Trained AlexNet model is used for computation of LPIPS distance. PGD-AT model based LPIPS distance is useful to distinguish between Oracle-Sensitive and Oracle-Invariant attacks.}
        \label{fig:lpips_plot}
        \vspace{0.4cm}
        \end{minipage}
        \hfill
        \begin{minipage}{0.48\linewidth}
\centering
        \includegraphics[width=\linewidth]{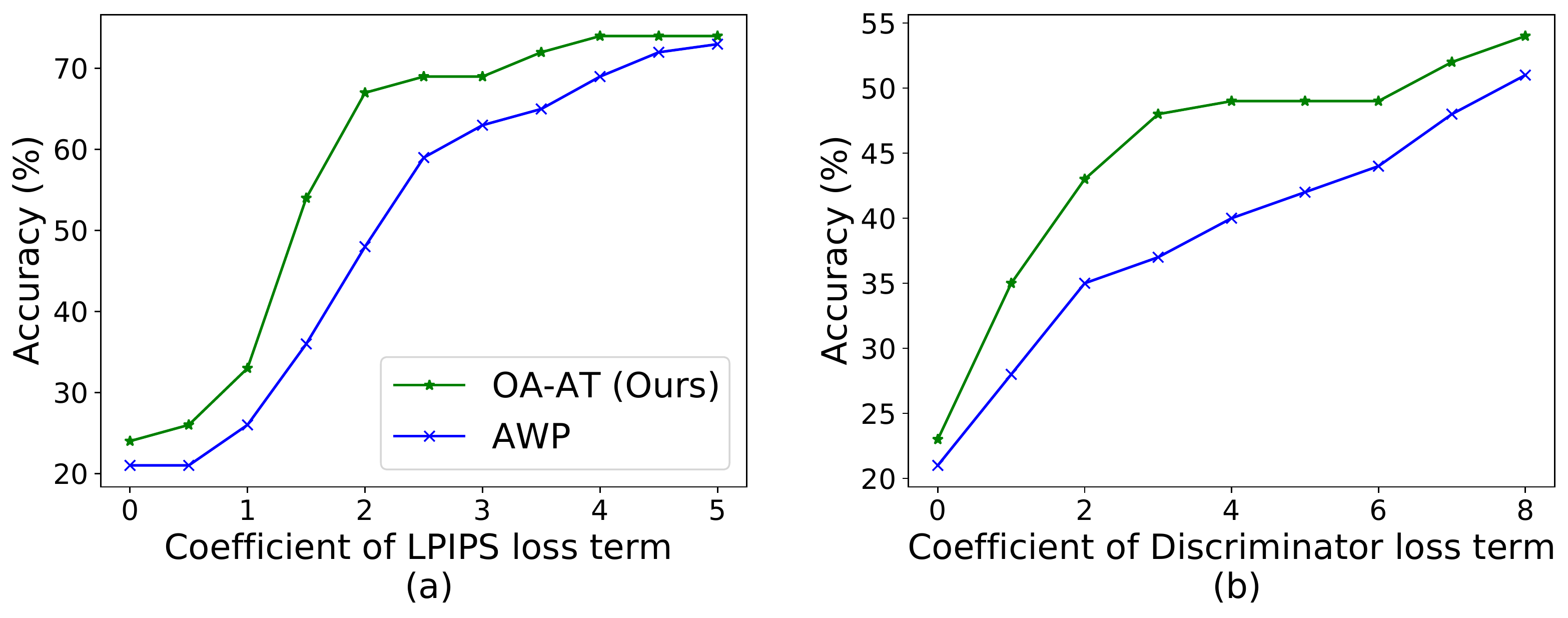}
        % \vspace{-0.4cm}
        \caption{Comparison of the proposed model with AWP \cite{wu2020adversarial} on CIFAR-10, against attacks of varying strength and Oracle sensitivity constrained within $\varepsilon=16/255$. (a) LPIPS based regularizer, (b) Discriminator based regularizer are used for generating Oracle-Invariant attacks respectively. As the coefficient of the regularizer increases, the attack transforms from Oracle-Sensitive to Oracle-Invariant. The proposed method achieves improved accuracy when compared to AWP.}
        \label{fig:sens_attack_eval}
        % \vspace{-0.5cm}
\end{minipage}
\vspace{-0.8cm}
\end{figure*}

\subsection{PGD based Attacks }While the most efficient attack that is widely used for adversarial training is the PGD $10$-step attack, it cannot be used for the generation of Oracle-Invariant samples as adversarially trained models have perceptually aligned gradients, and tend to produce Oracle-Sensitive samples. Therefore, we explore some variants of the PGD attack to make the generated perturbations Oracle-Invariant. We denote the Cross-Entropy loss on a data sample $x$ with ground truth label $y$ using $\mathcal{L}_{CE}(x,y)$. We explore the addition of regularizers to the Cross-Entropy loss weighted by a factor of $\lambda_{X}$ in each case. The value of $\lambda_{X}$ is chosen as the minimum value which transforms the PGD attacks from Oracle-Sensitive to Oracle-Invariant. This results in the strongest possible Oracle-Invariant attacks.

\subsection{Discriminator based PGD Attack} We train a discriminator to distinguish between Oracle-Invariant and Oracle-Sensitive adversarial examples, and further maximize the below loss for the generation of Oracle-Invariant attacks:

\begin{equation}
\label{eq:disc_app}
    \mathcal{L}_{CE}(x,y) - \lambda_{Disc} \cdot \mathcal{L}_{BCE}(\hat{x},\textrm{OI}) 
\end{equation}

Here $\mathcal{L}_{BCE}(\hat{x},\textrm{OI})$ is the Binary Cross-Entropy loss of the adversarial example $\hat{x}$ w.r.t. the label corresponding to an Oracle-Invariant (OI) attack. We train the discriminator to distinguish between two input distributions; the first corresponding to images concatenated channel-wise with their respective Oracle-Sensitive perturbations, and a second distribution where perturbations are shuffled across images in the batch. This ensures that the discriminator relies on the spatial correlation between the image and its corresponding perturbation for the classification task, rather than the properties of the perturbation itself. The attack in Eq.\ref{eq:disc_app} therefore attempts to break the most salient property of Oracle-Sensitive attacks, which is the spatial correlation between an image and its perturbation.

\begin{figure*}[t]
\centering
        \includegraphics[width=\linewidth]{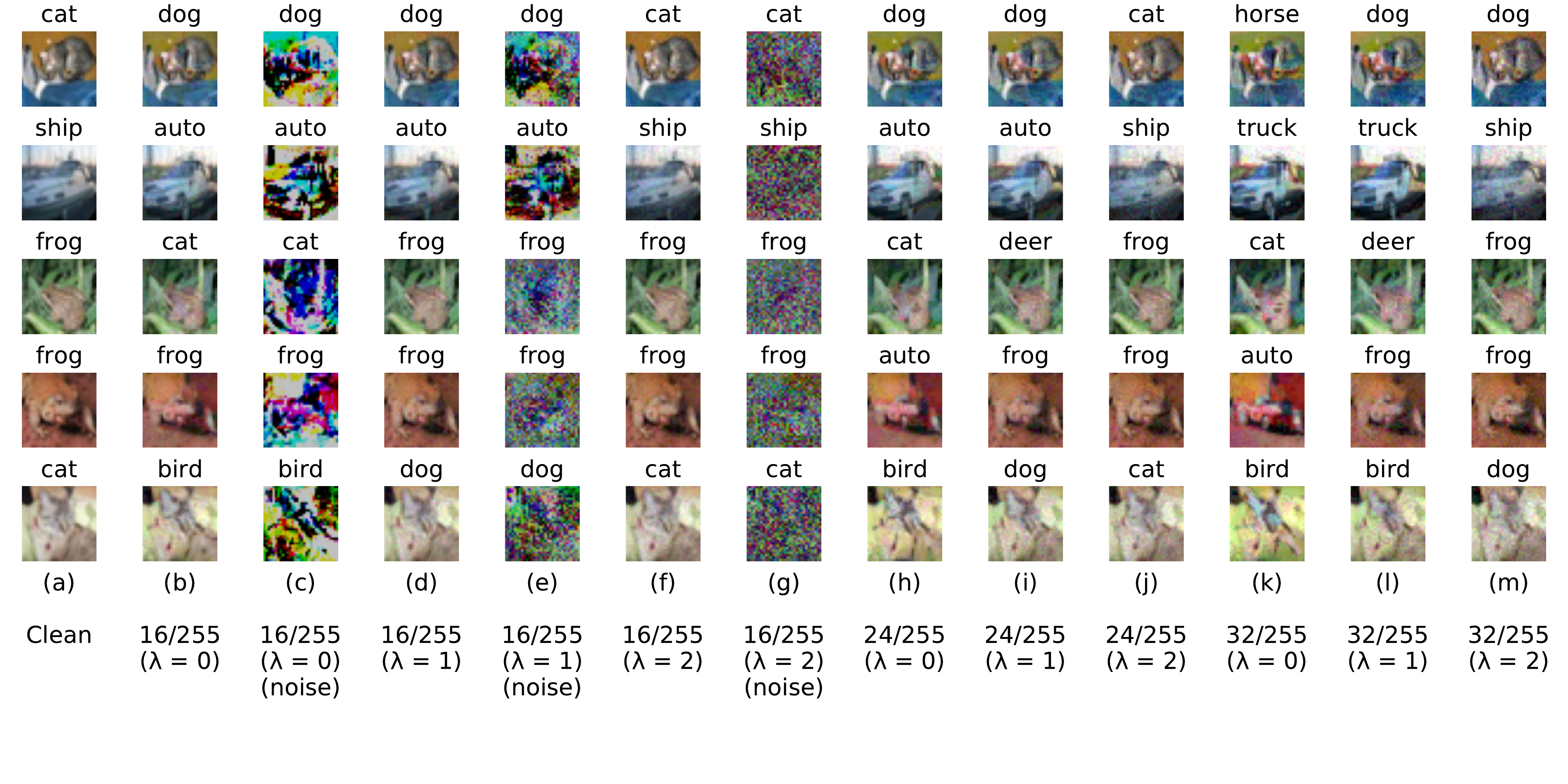}
        \vspace{-1.2cm}
        \caption{Oracle-Invariant adversarial examples generated using the LPIPS based PGD attack in Eq.\ref{eq:lpips_app} across various perturbation bounds. White-box attacks and predictions on the model trained using the proposed OA-AT defense on the CIFAR-10 dataset with ResNet-18 architecture are shown: (a) Original Unperturbed image, (b, h, k) Adversarial examples generated using the standard PGD 10-step attack, (d, f, i, j, l, m) LPIPS based PGD attack generated within perturbation bounds of $16/255$ (d, f), $24/255$ (i, j) and $32/255$ (l, m) by setting the value of $\lambda_\textrm{LPIPS}$ to 1 and 2, (c, e, g) Perturbations corresponding to (b), (d) and (f) respectively.}
        \label{fig:lpips_attack_app2}
        \vspace{-0.3cm}
\end{figure*}
\subsection{LPIPS based PGD Attack} We propose to use the Learned Perceptual Image Patch Similarity (LPIPS) measure for the generation of Oracle-Sensitive attacks, as it is known to match well with perceptual similarity \cite{zhang2018unreasonable,laidlaw2020perceptual}. As shown in Fig.\ref{fig:lpips_plot}, while the standard AlexNet model that is used in prior work \cite{laidlaw2020perceptual} fails to distinguish between Oracle-Invariant and Oracle-Sensitive samples, an adversarially trained model is able to distinguish between the two types of attacks effectively. In this plot, we consider attacks generated from a PGD-AT \cite{madry-iclr-2018,pang2020bag} model (Fig.\ref{fig:lp_pert}(c-e)) as Oracle-Sensitive attacks, and attacks generated from a Normally Trained model (Fig.\ref{fig:lp_pert}(h)) as Oracle-Invariant attacks. We therefore propose to minimize the LPIPS distance between the natural and perturbed images, in addition to the maximization of Cross-Entropy loss for attack generation as shown below:
\begin{equation}
\label{eq:lpips_app}
    \mathcal{L}_{CE}(x,y) - \lambda_{\textrm{LPIPS}} \cdot \textrm{LPIPS}(x,\hat{x})
\end{equation}

We choose $\lambda_{\textrm{LPIPS}}$ as the minimum value that transforms the PGD attack from Oracle-Sensitive to Oracle-Invariant (OI), to generate strong OI attacks. This is further fine-tuned during training to achieve the optimal robustness-accuracy trade-off. As shown in Fig.\ref{fig:lpips_attack_app2}, setting $\lambda_{\textrm{LPIPS}}$ to 1 changes adversarial examples from Oracle-Sensitive to Oracle-Invariant, as they look similar to the corresponding original images shown in Fig.\ref{fig:lpips_attack_app2} (a). This can be observed more distinctly at perturbation bounds of $24/255$ and $32/255$. The perturbations in Fig.\ref{fig:lpips_attack_app2} (c) are smooth, while those in (e) and (g) are not. This shows that the addition of the LPIPS term helps in making the perturbations Oracle-Invariant. Very large coefficients of the LPIPS term make the attack weak as can be seen in Fig.\ref{fig:lpips_attack_app2} (f, j, m) where the model prediction is same as the true label. We therefore set $\lambda_\textrm{LPIPS}$ to 1 to obtain strong Oracle-Invariant attacks. 
As shown in Table-\ref{table:ablations_new}, while we obtain the best results using the LPIPS based PGD attack for training (E1), the use of discriminator based PGD attack (E8) also results in a better robustness-accuracy trade-off when compared to E2, where there is no explicit regularizer to ensure the generation of Oracle-Invariant attacks.

\subsection{Evaluation of the proposed defense against Oracle-Invariant Attacks} 
We compare the performance of the proposed defense OA-AT with the strongest baseline AWP \cite{wu2020adversarial} against the two proposed Oracle-Invariant attacks, LPIPS based attack and Discriminator based attack in Fig.\ref{fig:sens_attack_eval} (a) and (b) respectively. We vary the coefficient of the regularizers used in the generation of attacks, $\lambda_{Disc}$ (Eq.\ref{eq:disc_app}) and $\lambda_{\textrm{LPIPS}}$ (Eq.\ref{eq:lpips_app}) in each of the plots. As we increase the coefficient, the attack transforms from Oracle-Sensitive to Oracle-Invariant. The proposed method (OA-AT) achieves improved accuracy compared to the AWP \cite{wu2020adversarial} baseline.

\section{Analysing Oracle Alignment of Adversarial Attacks}
\label{theorem_suppl}
In this section, we present more detailed analysis of generating Oracle-Invariant and Oracle-Sensitive attacks in a simplified yet natural setting. We consider a binary classification task as proposed by Tsipras et al. \cite{tsipras2018robustness}, consisting of data samples $(x,y)$, with $y\in\{+1,-1\}$, $x\in\mathbb{R}^{d+1}$. Further, 
$$x_1 = 
\begin{cases}
y,&\text{w.p. } p\\
-y,&\text{w.p. } 1-p\\
\end{cases}~,~ x_i \sim \mathcal{N}(\alpha y,1)~ \forall i \in \{2,\dots,d+1\} $$
In this setting, $x_1$ can be viewed as a feature that is strongly correlated with the Oracle Label $y$ when the Bernoulli parameter $p$ is sufficiently large (for eg: $p \approx 0.90$), and thus corresponds to an Oracle Sensitive feature.  On the other hand, $x_2,\dots,x_{d+1}$ are spurious features that are positively correlated (in a weak manner) to the Oracle label $y$, and are thus Oracle Invariant features. \\

Case 1: A simple, yet effective classifier that achieves high accuracy is given as follows: $f(x) := sign(w^Tx)$, where $w\in\mathbb{R}^{d+1}$ with $w=(0,\frac{1}{d},\dots,\frac{1}{d})$. Then, its accuracy is given by 
$$\mathbb{P}[f(x)=y] = \mathbb{P}[y\cdot sign(w^Tx)=1] = \mathbb{P}\left[y \sum_{i=2}^{d+1} \frac{x_i}{d} > 0\right] = \mathbb{P}[z>0]$$
where $z\sim\mathcal{N}(\alpha,\frac{1}{d})$. 

\noindent
We see this is true since $z$ is given by a sum of $d$ i.i.d. Gaussian random variables $y x_i/d$, each with mean $y\cdot \alpha y/d = \alpha /d$ (since $y^2 =1$), and with variance $1/d^2$ each. Thus the accuracy exceeds $99\%$ if $\alpha>\frac{3}{\sqrt{d}}$ by properties of the Gaussian distribution. We note that such a classifier that achieves vanishing error can be learnt through standard Empirical Risk Minimisation.

However, with an $\ell_{\infty}$ attack with perturbation budget $2\alpha$, an adversary can flip each of the weakly correlated features $x_2,\dots,x_{d+1}$ to appear as $x_i+\delta \sim \mathcal{N}(-\alpha y,1)$. These perturbed features are thus now weakly anti-correlated with the Oracle label $y$, and achieves a robust accuracy of less than 1\%. Thus by attacking a standard, non-robust classifier the perturbations can be seen to be Oracle Invariant features that are spurious in nature.
Tsipras et al. \cite{tsipras2018robustness} prove a more general version of the same result:

\vspace{0.2cm}
\noindent
\textbf{Theorem [Tsipras et al.]} Let $f$ be any classifier that achieves standard accuracy of at least $1-\gamma$, that is $\mathbb{P}[f(x)=y]>1-\gamma$. Then, the robust accuracy achieved by $f$ under an $\ell_{\infty}$ attack with a perturbation budget of $2\alpha$, has a tight upper-bound given by $\frac{p}{1-p}\gamma$. 

\vspace{0.2cm}
\noindent
\textbf{Observation 1.} Adversarial perturbations of a standard, non-robust classifier utilize spurious features, resulting in Oracle Invariant Samples that are weakly anti-correlated with the Oracle label $y$.

\vspace{0.4cm}
Case 2: Here, we consider another simple classifier, $g(x) := sign(x_1)$, which achieves, on natural samples, an accuracy of $$\mathbb{P}[g(x)=y] = \mathbb{P}[y\cdot sign(x_1) = 1] = p$$
While the accuracy thus has a tight upper-bound of $p$, the classifier $g$ is robust to adversarial perturbations of relatively large magnitude. Further, to maximise the misclassification of model $g$, it is easy to see that adversarial perturbations take the form $\delta=(2\alpha,0,\dots,0)$. Thus, we observe that adversarial perturbations of robust models correspond to Oracle Sensitive features.

\vspace{0.2cm}
\noindent
\textbf{Observation 2.} Adversarial perturbations of a robust model result in Oracle Sensitive Samples, utilizing features strongly correlated with the Oracle label $y$.

\begin{theorem} 
Consider a robust Deep Neural Network $f_{\theta}$ with parameters $\theta$ as in Algorithm-1. Given an input sample $x$, let $\delta^*$ represent an optimal solution that maximises the following objective:
\vspace{-0.1cm}
\begin{equation}
\ell = \ell_{CE}(f_{\theta}(x+\delta),y)  - \lambda \cdot \LPIPS(x,x+\delta) 
\end{equation}
\vspace{-0.2cm}
\noindent
Then, $\exists ~\lambda>0$ such that $x+\delta^*$ is an Oracle Invariant Sample.
\end{theorem}

\noindent
\textbf{Proof:} By definition, the LPIPS metric between samples $x$ and $x+\delta^*$ measures aggregate L2 distances between the corresponding feature space representations in the intermediate layers of robust network $f_{\theta}$. For $\lambda>>0$, the LPIPS component dominates the overall optimization objective in the adversarial attack. To prove the result, let us assume on the contrary that the perturbation $\delta^*$ results in an Oracle Sensitive Sample. Thus the corresponding feature representations in a robust network for the sample $x+\delta^*$ would deviate significantly from that of the original benign sample $x$. Thus, as $\LPIPS(x,x+\delta^*)>0$, and  as $\lambda 	\rightarrow \infty$, the overall objective in Eqn(1) decreases, with $\ell\rightarrow -\infty$, contradicting the optimality of $\delta^*$ in maximising the same objective. Thus, we conclude that $x+\delta^*$ is indeed an Oracle Invariant Sample.

\section{Details on the Datasets used}
We evaluate the proposed approach on the CIFAR-10, CIFAR-100 \cite{krizhevsky2009learning} and SVHN \cite{svhn} datasets. The three datasets consist of RGB images of spatial dimension 32$\times$32. CIFAR-10 and SVHN  contain $10$ distinct classes, while CIFAR-100 contains $100$. CIFAR-10 is the most widely used benchmark dataset to perform a comparative analysis across different adversarial defense and attack methods. CIFAR-100 is a challenging dataset to achieve adversarial robustness given the large number of diverse classes that are interrelated. Each of these datasets consists of 50,000 training images and 10,000 test images, while SVHN contains 73257 training and 26032 testing images. We split the original training set to create a validation set of 1,000 images in CIFAR-10 and 2,500 images in CIFAR-100 and SVHN. We ensure that the validation split is balanced equally across all classes, and use the remaining images for training. To ensure a fair comparison, we use the same split for training the proposed defense as well as other baseline approaches. For CIFAR-10 and CIFAR-100 datasets, we consider the $\ell_{\infty}$ threat model of radius $8/255$ to be representative of imperceptible perturbations, that is, the Oracle label does not change within this set. For SVHN we consider this bound to be $4/255$ as many of the images in the dataset have a low contrast, leading to visible perturbations at relatively small $\varepsilon$ bounds.  Further, to investigate robustness within moderate magnitude perturbation bounds, we consider the $\ell_{\infty}$ threat model of radius $16/255$  for CIFAR-10 and CIFAR-100, and a bound of $12/255$ for SVHN.

\section{Details on Training}

In this section, we expound further details on the algorithm of the proposed method, presented in Sec.\ref{sec:prop_meth} (Alg.\ref{alg:train_algo}). We use a varying $\varepsilon$ schedule and start training on perturbations of magnitude $\varepsilon_{max}/4$. This results in marginally better performance when compared to ramping up the value of $\varepsilon$ from 0 (E9 of Table-\ref{table:ablations_new}). 
For CIFAR-10 training on ResNet-18, we set the weight of the adversarial loss $L_{adv}$ in L21 of Alg.1 ($\beta$ parameter of TRADES \cite{zhang2019theoretically}) to 1.5 for the first three-quarters of training, and then linearly increase it from 1.5 to 3 in the moderate perturbation regime, where $\varepsilon$ is linearly increased from $12/255$ to $16/255$. In this moderate perturbation regime, we also linearly increase the coefficient of the LPIPS distance (Alg.1, L14) from 0 to 1, and linearly decrease the $\alpha$ parameter used in the convex combination of softmax prediction (Alg.1, L11) from 1 to 0.8. This results in a smooth transition from adversarial training on imperceptible attacks to attacks with larger perturbation bounds. We set the weight decay to 5e-4.

We use cosine learning rate schedule with a maximum learning rate of 0.2 for CIFAR-10 and CIFAR-100, and 0.05 for SVHN. We use SGD optimizer with momentum of 0.9, and train for 110 epochs, except for training PreActResNet18 on CIFAR-100 and WideResNet-34-10 on CIFAR-10,  where we use 200 epochs. We do not perform early stopping and always report accuracy of the last epoch. We compute the LPIPS distance using an exponential weight averaged model with $\tau=0.995$. We note from Table-\ref{table:ablations_new} that the use of weight-averaged model (E1) results in better performance when compared to using the model being trained for the same (E5). This also leads to more stable results across reruns. 

\begin{table}[t]
\caption{\textbf{CIFAR-10, CIFAR-100}: Ablation experiments on ResNet-18 architecture to highlight the importance of various aspects in the proposed defense OA-AT. Performance ($\%$) against attacks with different perturbation bounds $\varepsilon$ is reported.}
\setlength\tabcolsep{3pt}
\resizebox{1.0\linewidth}{!}{
\label{table:ablations_new}
\begin{tabular}{l|c|c|cc||c|c|cc}
\toprule
                                                              & \multicolumn{4}{c||}{\textbf{CIFAR-10}}                                                                                                                                                                                           & \multicolumn{4}{c}{\textbf{CIFAR-100}}                                                                                                                                                                                    \\
\cmidrule(l){2-9}
\textbf{Method}                                               & \textbf{~Clean~}       & \textbf{\begin{tabular}[c]{@{}c@{}}~GAMA~\\ (8/255)\end{tabular}} & \textbf{\begin{tabular}[c]{@{}c@{}}~GAMA~\\ (16/255)\end{tabular}} & \textbf{\begin{tabular}[c]{@{}c@{}}~Square~ \\ (16/255)\end{tabular}} & \textbf{~Clean~} & \textbf{\begin{tabular}[c]{@{}c@{}}~GAMA~\\ (8/255)\end{tabular}} & \textbf{\begin{tabular}[c]{@{}c@{}}~GAMA~\\ (16/255)\end{tabular}} & \textbf{\begin{tabular}[c]{@{}c@{}}~Square~ \\ (16/255)\end{tabular}} \\
\midrule
\textbf{E1}: OA-AT (Ours)                                      & 80.24                & \textbf{51.40}                                                  & 22.73                                                            & 31.16                                                               & 60.27          & \textbf{26.41}                                                  & 10.47                                                            & 14.60                                                               \\
\textbf{E2}: LPIPS weight = 0                                 & 78.47                & 50.60                                                           & 24.05                                                            & 31.37                                                               & 58.47          & 25.94                                                           & 10.91                                                            & 14.66                                                               \\
\textbf{E3}: Alpha = 1                                        & 79.29                & 50.60                                                           & 23.65                                                            & 31.23                                                               & 58.84          & 26.15                                                           & 10.97                                                            & 14.89                                                               \\
\textbf{E4}: Alpha = 1, LPIPS weight = 0                      & 77.16                & 50.49                                                           & \textbf{24.93}                                                   & \textbf{32.01}                                                      & 57.77          & 25.92                                                           & \textbf{11.33}                                                   & \textbf{15.03}                                                      \\
\textbf{E5}: Using Current model (without WA) for LPIPS       & 80.50                & 50.75                                                           & 22.90                                                            & 30.76                                                               & 59.54          & 26.23                                                           & 10.50                                                            & 14.86                                                               \\

\textbf{E6}: Without 2*eps perturbations for AWP              & 79.96                & 50.50                                                           & 22.61                                                            & 30.60                                                               & 60.18          & 26.27                                                           & 10.15                                                            & 14.20                                                               \\

\textbf{E7}: Maximizing KL div in the AWP step                & 81.19                & 49.77                                                           & 21.17                                                            & 29.39                                                               & 59.48          & 25.03                                                           & 7.93                                                             & 13.34                                                               \\
\textbf{E8}: Using Discriminator instead of LPIPS (OI Attack)~~ & 80.56                & 50.75                                                           & 22.13                                                            & 31.17                                                               & 58.84          & 26.35                                                           & 10.64                                                            & 14.82                                                               \\
\textbf{E9}: Increasing epsilon from the beginning            & 80.34                & 50.77                                                           & 22.57                                                            & 30.80                                                               & \textbf{60.51} & 26.34                                                           & 10.37                                                            & 14.61                                                               \\
\textbf{E10}: Without AutoAugment                             & 80.24                & \textbf{51.40}                                                           & 22.73                                                            & 31.16                                                               & 58.08          & 25.81                                                           & 10.40                                                            & 14.31                                                               \\
\textbf{E11}: With AutoAugment (p=0.5)                        & 81.59       & 50.40                                                           & 21.59                                                            & 30.84                                                               & 60.27          & \textbf{26.41}                                                           & 10.47                                                            & 14.60                                                               \\
\textbf{E12}: With AutoAugment (p=1) & \textbf{81.74}&	48.15&	18.92&	28.31& 60.19 & 25.32 & 9.24 & 13.78  \\
\textbf{E13}: Alpha = 1, LPIPS weight = 0 + fixed $\varepsilon$=16/255 & 71.64&	47.59&	25.91&	31.75& 50.99 & 23.19 & 9.99 & 13.48  \\
\bottomrule
\end{tabular}}
\vspace{-0.3cm}
\end{table}

We utilise AutoAugment \cite{cubuk2018autoaugment} for training on CIFAR-100, SVHN and for CIFAR-10 training on large model capacities. We apply AutoAugment with a probability of $0.5$ for CIFAR-100, and for the CIFAR-10 model trained on ResNet-34. Since the extent of overfitting is higher for large model capacities, we use AutoAugment with $p=1$ on WideResNet-34-10. While the use of AutoAugment helps in overcoming overfitting, it could also negatively impact robust accuracy due to the drift between the training and test distributions. We observe a drop in robust accuracy on the CIFAR-10 dataset with the use of AutoAugment (E11, E12 in Table-\ref{table:ablations_new}), while there is a boost in the clean accuracy. On similar lines, we observe a drop in robust accuracy on the CIFAR-100 dataset as well, when we increase the probability of applying AutoAugment from $0.5$ (E11 in Table-\ref{table:ablations_new}) to 1 (E12 in Table-\ref{table:ablations_new}). 
We use AutoAugment with $p=1$ for SVHN, as we observe that it results in more stable training. Further, we find that using Label Smoothing with CIFAR-100 helps in improving clean accuracy, as shown in Table-\ref{table:cifar10_main}.

To investigate the stability of the proposed approach, we train a ResNet-18 network multiple times by using different random initialization of
network parameters. We observe that the proposed approach is indeed stable, with standard deviation of $0.167$, $0.115$, $0.180$ and $0.143$ for clean accuracy, GAMA PGD-100 accuracies with $\varepsilon = 8/255$ and $16/255$, and accuracy against the Square attack with $\varepsilon = 16/255$ respectively over three independent training runs on CIFAR-10. We also observe that the last epoch is consistently the best performing model for the ResNet-18 architecture. Nonetheless, we still utilise early stopping on the validation set using PGD 7-step accuracy for all the baselines to enable a fair comparison overall. 

\section{Evaluation of Adversarial Defenses}
\label{adaptive_supp}
Gradient-based white-box attacks such as PGD \cite{madry-iclr-2018}, GAMA-PGD \cite{sriramanan2020gama} and Auto-PGD with Cross-Entropy (CE) and Difference of Logits Ratio (DLR) losses \cite{croce2019minimally} are known to be the strongest attacks against standard Adversarial defenses that do not obfuscate gradients. Gradient-Free attacks such as ZOO \cite{chen2017zoo}, SPSA \cite{uesato2018adversarial}, Square \cite{andriushchenko2019square} and RayS \cite{chen2020rays} are useful to craft perturbations without requiring white-box access to the model. These attacks are also used to reliably estimate the robustness of defenses that rely on gradient masking \cite{papernot2017practical}. 
Amongst the Gradient-Free attacks, Square and Ray-S do not use Zeroth order gradient estimates, and utilize Random-Search and Binary-Search based algorithms respectively to construct strong attacks against a given defense. We use such query-based attacks to generate perturbations that do not flip Oracle predictions even for moderate-magnitude constraint sets. AutoAttack combines strong untargeted and targeted white-box attacks with the query-based black-box attack Square to effectively estimate the robustness of a given defense, and is a well accepted standard for benchmarking defenses. We report our results against GAMA-PGD, AutoAttack, Square and Ray-S. We also present further evaluations using various adaptive attacks (Sec.\ref{adaptive_supp}) to reliably estimate robustness of the proposed defense.

\section{Ablation Study}
\label{ablations_supp}
In order to study the impact of different components of the proposed defense, we present a detailed ablative study using ResNet-18 models in Table-\ref{table:ablations_new}. We present results on the CIFAR-10 and CIFAR-100 datasets, with E1 representing the proposed approach. First, we study the efficacy of the LPIPS metric in generating Oracle-Invariant attacks. In experiment E2, we train a model without LPIPS by setting its coefficient to zero. While the resulting model achieves a slight boost in robust accuracy at $\varepsilon=16/255$ due to the use of stronger attacks for training, there is a considerable drop in clean accuracy, and a corresponding drop in robust accuracy at $\varepsilon=8/255$ as well. We observe a similar trend by setting the value of $\alpha$ to $1$ as shown in E3, and by combining E2 and E3 as shown in E4. We note that E4 is similar to standard adversarial training, where the model attempts to learn consistent predictions in the $\varepsilon$ ball around every data sample. While this works well for large $\varepsilon$ attacks ($\varepsilon=16/255$), it leads to poor clean accuracy as shown in the first partition of Table-\ref{table:cifar10_full}.

As discussed in Sec.\ref{sec:prop_meth}, we maximize loss on $x_i+2\cdot\widetilde{\delta_i}$ (where $\widetilde{\delta_i}$ is the attack) in the additional weight perturbation step. We present results by using the standard $\varepsilon$ limit for the weight perturbation step as well, in E6. This leads to a drop across all metrics, indicating the importance of using large magnitude perturbations in the weight perturbation step for producing a flatter loss surface that leads to better generalization to the test set. Different from the standard TRADES formulation, we maximize Cross-Entropy loss for attack generation in the proposed method. From E7, we note that the use of KL divergence leads to a drop in robust accuracy since the KL divergence based attack is weaker. This is consistent with the observation by Gowal et al. \cite{gowal2020uncovering}. However, on the SVHN dataset, we find that the use of KL divergence based attack results in a significant improvement in clean accuracy, leading to better robust accuracy as well. We therefore utilize the KL divergence loss for attack generation on the SVHN dataset.  We also investigate the effect of AutoAugment \cite{cubuk2018autoaugment}, Weight Averaging \cite{izmailov2018averaging} and Label Smoothing + Warmup on the AWP \cite{wu2020adversarial} baseline in Table-\ref{table:awp_plus}.

\section{Detailed Results}
\label{detailed_res_supp}

\begin{table*}[t]
\caption{\textbf{CIFAR-10}: Performance ($\%$) of the proposed defense OA-AT against attacks bounded within different $\varepsilon$ bounds, when compared to the following baselines: AWP \cite{wu2020adversarial}, ExAT \cite{shaeiri2020towards}, TRADES \cite{zhang2019theoretically}, ATES \cite{sitawarin2020improving}, PGD-AT \cite{madry-iclr-2018} and FAT \cite{zhang2020attacks}. AWP \cite{wu2020adversarial} is the strongest baseline. The first partition shows defenses trained on $\varepsilon=16/255$. Training on large perturbation bounds results in very poor Clean Accuracy. The second partition consists of baselines tuned to achieve clean accuracy close to $80\%$. These are sorted by AutoAttack accuracy \cite{croce2020reliable} (AA $8/255$). The proposed defense (OA-AT) achieves significant gains in accuracy across all attacks.}
\vspace{0.1cm}
\setlength\tabcolsep{1pt}
\resizebox{1.0\linewidth}{!}{
\label{table:cifar10_full}
\begin{tabular}{l|c|c|cccc|cccc|cccc}
\toprule
\textbf{Method}              & \textbf{\begin{tabular}[c]{@{}c@{}}Attack $\varepsilon$ \\ \small{(Training)}\end{tabular}} & \textbf{Clean} & \textbf{\begin{tabular}[c]{@{}c@{}}FGSM (BB)\\ (8/255)\end{tabular}} & \textbf{\begin{tabular}[c]{@{}c@{}}R-FGSM \\ (8/255)\end{tabular}} & \textbf{\begin{tabular}[c]{@{}c@{}}GAMA \\ (8/255)\end{tabular}} & \textbf{\begin{tabular}[c]{@{}c@{}}AA \\ (8/255)\end{tabular}} & \textbf{\begin{tabular}[c]{@{}c@{}}FGSM (BB)\\ (12/255)\end{tabular}} & \textbf{\begin{tabular}[c]{@{}c@{}}R-FGSM \\ (12/255)\end{tabular}} & \textbf{\begin{tabular}[c]{@{}c@{}}GAMA \\ (12/255)\end{tabular}} & \textbf{\begin{tabular}[c]{@{}c@{}}Square \\ (12/255)\end{tabular}} & \textbf{\begin{tabular}[c]{@{}c@{}}FGSM (BB)\\ (16/255)\end{tabular}} & \textbf{\begin{tabular}[c]{@{}c@{}}R-FGSM \\ (16/255)\end{tabular}} & \textbf{\begin{tabular}[c]{@{}c@{}}GAMA \\ (16/255)\end{tabular}} & \textbf{\begin{tabular}[c]{@{}c@{}}Square \\ (16/255)\end{tabular}} \\
\midrule
TRADES                       & 16/255                                                             & 75.30          & 73.26                                                                & 53.10                                                              & 35.64                                                                & 35.12                                                          & 72.13                                                                 & 44.27                                                               & 20.24                                                                 & 30.11                                                               & 70.76                                                                 & 36.99                                                               & 10.10                                                                 & 18.87                                                               \\
AWP                          & 16/255                                                             & 71.63          & 69.71                                                                & 54.53                                                              & 40.85                                                                & 40.55                                                          & 68.65                                                                 & 47.13                                                               & 27.06                                                                 & 34.42                                                               & 67.42                                                                 & 40.89                                                               & 15.92                                                                 & 24.16                                                               \\
PGD-AT                       & 16/255                                                             & 64.93          & 63.65                                                                & 55.47                                                              & 46.66                                                                & 46.21                                                          & 62.81                                                                 & 51.05                                                               & 36.95                                                                 & 40.53                                                               & 61.70                                                                 & 46.40                                                               & 26.73                                                                 & 32.25                                                               \\
FAT                          & 16/255                                                             & 75.27          & 73.44                                                                & 60.25                                                              & 47.68                                                                & 47.34                                                          & 72.22                                                                 & 53.17                                                               & 34.31                                                                 & 39.79                                                               & 70.73                                                                 & 46.88                                                               & 22.93                                                                 & 29.47                                                               \\
ExAT+AWP                     & 16/255                                                             & 75.28          & 73.27                                                                & 60.02                                                              & 47.63                                                                & 47.46                                                          & 71.81                                                                 & 52.38                                                               & 34.42                                                                 & 39.62                                                               & 70.47                                                                 & 45.39                                                               & 22.61                                                                 & 28.79                                                               \\
ATES & 16/255                                     & 66.78          & 65.60                                                                 & 56.79                                                              & 47.89                                                                & 47.52                                                          & 64.64                                                                 & 51.71                                                               & 37.47                                                                 & 42.07                                                               & 63.75                                                                 & 47.28                                                               & 26.50                                                                  & 32.55                                                               \\
ExAT + PGD                   & 16/255                                                             & 72.04          & 70.68                                                                & 59.99                                                              & 49.24                                                                & 48.80                                                          & 69.66                                                                 & 53.96                                                               & 36.68                                                                 & 41.93                                                               & 68.04                                                                 & 48.37                                                               & 23.01                                                                 & 30.21                                                               \\
\midrule

FAT                          & 12/255                                                             & 80.27          & 77.87                                                                & 61.46                                                              & 45.42                                                                & 45.13                                                          & 76.69                                                                 & 52.33                                                               & 29.08                                                                 & 36.71                                                               & 74.79                                                                 & 44.56                                                               & 16.18                                                                 & 24.59                                                               \\
FAT                          & 8/255                                                              & \textbf{84.36} & 82.20                                                                & 64.06                                                              & 48.41                                                                & 48.14                                                          & 80.32                                                                 & 55.41                                                               & 29.39                                                                 & 39.48                                                               & 78.13                                                                 & 47.50                                                               & 15.18                                                                 & 25.07                                                               \\
ATES                         & 8/255                                                              & 84.29          & \textbf{82.39}                                                       & \textbf{65.66}                                                     & 49.14                                                                & 48.56                                                          & \textbf{80.81}                                                        & 55.59                                                               & 29.36                                                                 & 40.68                                                               & \textbf{78.48}                                                        & 47.03                                                               & 14.70                                                                 & 25.88                                                               \\
PGD-AT                       & 8/255                                                              & 81.12          & 78.94                                                                & 63.48                                                              & 49.03                                                                & 48.58                                                          & 77.19                                                                 & 54.42                                                               & 30.84                                                                 & 40.82                                                               & 74.37                                                                 & 46.28                                                               & 15.77                                                                 & 26.47                                                               \\
PGD-AT                       & 10/255                                                             & 79.38          & 77.89                                                                & 62.78                                                              & 49.28                                                                & 48.68                                                          & 76.60                                                                 & 54.76                                                               & 32.40                                                                 & 41.46                                                               & 74.75                                                                 & 47.46                                                               & 18.18                                                                 & 28.29                                                               \\
AWP                          & 10/255                                                             & 80.32          & 77.87                                                                & 62.33                                                              & 49.06                                                                & 48.89                                                          & 76.33                                                                 & 53.83                                                               & 32.88                                                                 & 40.27                                                               & 74.13                                                                 & 45.51                                                               & 19.17                                                                 & 27.56                                                               \\
ATES                         & 10/255                                                             & 80.95          & 79.22                                                                & 63.95                                                              & 49.57                                                                & 49.12                                                          & 77.77                                                                 & 55.37                                                               & 32.44                                                                 & 42.21                                                               & 75.51                                                                 & 48.12                                                               & 18.36                                                                 & 29.07                                                               \\

TRADES                       & 8/255                                                              & 80.53          & 78.58                                                                & 63.69                                                              & 49.63                                                                & 49.42                                                          & 77.20                                                                 & 55.48                                                               & 33.32                                                                 & 40.94                                                               & 75.05                                                                 & 47.92                                                               & 19.27                                                                 & 27.82                                                               \\
ExAT + PGD                   & 11/255                                                             & 80.68          & 79.07                                                                & 63.58                                                              & 50.06                                                                & 49.52                                                          & 77.98                                                                 & 55.92                                                               & 32.47                                                                 & 41.10                                                               & 76.12                                                                 & 48.37                                                               & 17.81                                                                 & 27.23                                                               \\
ExAT + AWP                   & 10/255                                                             & 80.18          & 78.04                                                                & 63.15                                                              & 49.87                                                                & 49.69                                                          & 76.34                                                                 & 54.64                                                               & 33.51                                                                 & 41.04                                                               & 74.37                                                                 & 46.54                                                               & 20.04                                                                 & 28.40                                                               \\
AWP                          & 8/255                                                              & 80.47          & 78.22                                                                & 63.32                                                              & 50.06                                                                & 49.87                                                          & 76.88                                                                 & 54.61                                                               & 33.47                                                                 & 41.05                                                               & 74.42                                                                 & 46.16                                                               & 19.66                                                                 & 28.51                                                               \\

\textbf{OA-AT (Ours)}        & 16/255                                                             & 80.24          & 78.54                                                                & 65.00                                                              & \textbf{51.40}                                                       & \textbf{50.88}                                                 & 77.34                                                                 & \textbf{57.68}                                                      & \textbf{36.01}                                                        & \textbf{43.20}                                                      & 75.72                                                                 & \textbf{51.13}                                                      & \textbf{22.73}                                                        & \textbf{31.16}     \\ 
\textcolor{ao}{$^{\textrm{~~~~~~~~Gain w.r.t. AWP}}$}  & & \textcolor{gray}{$^{~~-0.23}$}         & \textcolor{ao}{$^{~~+0.32}$}                                                                 & \textcolor{ao}{$^{~~+1.68}$}                                                               & \textcolor{ao}{$^{~~+1.34}$}                                                                 & \textcolor{ao}{$^{~~+1.01}$}                                                           & \textcolor{ao}{$^{~~+0.46}$}                                                                  & \textcolor{ao}{$^{~~+3.07}$}                                                                & \textcolor{ao}{$^{~~+2.54}$}                                                                  & \textcolor{ao}{$^{~~+2.15}$}                                                                & \textcolor{ao}{$^{~~+1.30}$}                                                                  & \textcolor{ao}{$^{~~+4.97}$}                                                                & \textcolor{ao}{$^{~~+3.07}$}                                                                  & \textcolor{ao}{$^{~~+2.65}$}                                                                \\

\bottomrule
\end{tabular}}
\end{table*}
In Tables-\ref{table:cifar10_full} and  \ref{table:cifar100_full}, we present results of different defense methods such as AWP-TRADES \cite{wu2020adversarial}, TRADES \cite{zhang2019theoretically}, PGD-AT \cite{madry-iclr-2018}, ExAT \cite{shaeiri2020towards}, ATES \cite{sitawarin2020improving} and FAT \cite{zhang2020attacks}, evaluated across a wide range of adversarial attacks. We present evaluations on the Black-Box FGSM attack \cite{goodfellow2014explaining} and a suite of White-Box attacks, on $\ell_{\infty}$ constraint sets of different radii: $8/255$, $12/255$ and $16/255$. The white-box evaluations consist of the single-step Randomized-FGSM (R-FGSM) attack \cite{tramer2017ensemble}, the GAMA PGD-100 attack \cite{sriramanan2020gama} and AutoAttack \cite{croce2020reliable}, with the latter two being amongst the strongest of attacks known to date. Lastly, we also present evaluations on the Square attack \cite{andriushchenko2019square} for $\varepsilon=12/255$ and $16/255$ in order to evaluate performance on Oracle-Invariant samples at large perturbation bounds.

\subsection{CIFAR-10} To enable a fair comparison of the proposed approach with existing methods, we present comprehensive results of various defenses trained with different attack strengths in Table-\ref{table:cifar10_full}. In the first partition of the table, we present baselines trained using attacks constrained within an $\ell_{\infty}$ bound of $16/255$. While these models do achieve competitive robustness on adversaries of attack strength $\varepsilon = 8/255$, $12/255$ and $16/255$, they achieve significantly lower accuracy on clean samples which limits their use in practical scenarios. Thus, for better comparative analysis that accounts for the robustness-accuracy trade-off, we present results of the existing methods with hyperparameters and attack strengths tuned to achieve the best robust performance, while maintaining clean accuracy close to $80\%$ as commonly observed on the CIFAR-10 dataset on ResNet-18 architecture, in the second partition of Table-\ref{table:cifar10_full}. We observe that the proposed method OA-AT consistently outperforms other approaches on all three metrics described in Sec.\ref{sec:obj_pd}, by afchieving enhanced performance at $\varepsilon=8/255$ and $16/255$, while striking a favourable robustness-accuracy trade-off as well. The proposed defense achieves better robust performance even on the standard $\ell_{\infty}$ constraint set of $8/255$ when compared to existing approaches, despite being trained on larger perturbations sets.

\subsection{CIFAR-100} In Table-\ref{table:cifar100_full}, we present results on models trained on the highly-challenging CIFAR-100 dataset. Since this dataset contains relatively fewer training images per class, we seek to enhance performance further by incorporating the augmentation technique, AutoAugment \cite{cubuk2018autoaugment,stutz2021relating}. To enable fair comparison, we incorporate AutoAugment for the strongest baseline, AWP \cite{wu2020adversarial} as well. We observe that the proposed method consistently performs better than existing approaches by significant margins, both in terms of clean accuracy, as well as robustness against adversarial attacks conforming to the three distinct constraint sets. Further, this also confirms that the proposed method scales well to large, complex datasets, while maintaining a consistent advantage in performance compared to other approaches.

\begin{table*}[t]
\caption{\textbf{CIFAR-100}: Performance ($\%$) of the proposed defense OA-AT against attacks bounded within different $\varepsilon$ bounds, when compared to the following baselines: AWP \cite{wu2020adversarial}, ExAT \cite{shaeiri2020towards}, TRADES \cite{zhang2019theoretically}, ATES \cite{sitawarin2020improving}, PGD-AT \cite{madry-iclr-2018} and FAT \cite{zhang2020attacks}. AWP \cite{wu2020adversarial} is the strongest baseline. The baselines are sorted by AutoAttack accuracy \cite{croce2020reliable} (AA $8/255$). The proposed defense achieves significant gains in accuracy against the strongest attacks across all $\varepsilon$ bounds. Since the proposed defense uses AutoAugment \cite{cubuk2018autoaugment} as the augmentation strategy, we present results on the strongest baseline AWP \cite{wu2020adversarial} with AutoAugment as well.}
\vspace{0.1cm}
\setlength\tabcolsep{1pt}
\resizebox{1.0\linewidth}{!}{
\label{table:cifar100_full}
\begin{tabular}{l|c|c|cccc|cccc|cccc}
\toprule
\textbf{Method}              & \textbf{\begin{tabular}[c]{@{}c@{}}Attack $\varepsilon$ \\ \small{(Train)}\end{tabular}} & \textbf{Clean} & \textbf{\begin{tabular}[c]{@{}c@{}}FGSM-BB\\ (8/255)\end{tabular}} & \textbf{\begin{tabular}[c]{@{}c@{}}R-FGSM \\ (8/255)\end{tabular}} & \textbf{\begin{tabular}[c]{@{}c@{}}GAMA \\ (8/255)\end{tabular}} & \textbf{\begin{tabular}[c]{@{}c@{}}AA \\ (8/255)\end{tabular}} & \textbf{\begin{tabular}[c]{@{}c@{}}FGSM-BB\\ (12/255)\end{tabular}} & \textbf{\begin{tabular}[c]{@{}c@{}}R-FGSM \\ (12/255)\end{tabular}} & \textbf{\begin{tabular}[c]{@{}c@{}}GAMA \\ (12/255)\end{tabular}} & \textbf{\begin{tabular}[c]{@{}c@{}}Square \\ (12/255)\end{tabular}} & \textbf{\begin{tabular}[c]{@{}c@{}}FGSM-BB\\ (16/255)\end{tabular}} & \textbf{\begin{tabular}[c]{@{}c@{}}R-FGSM \\ (16/255)\end{tabular}} & \textbf{\begin{tabular}[c]{@{}c@{}}GAMA \\ (16/255)\end{tabular}} & \textbf{\begin{tabular}[c]{@{}c@{}}Square \\ (16/255)\end{tabular}} \\
\midrule
FAT                                   & 8/255                & 56.61                    & 52.10                     & 34.76                    & 23.36                    & 23.20                    & 49.54                     & 27.77                    & 13.96                    & 18.21                    & 46.01                     & 22.52                    & 8.30                     & 11.56                    \\
TRADES                                & 8/255                & 58.27                    & 54.33                     & 36.20                    & 23.67                    & 23.47                    & 51.64                     & 28.55                    & 13.88                    & 18.46                    & 48.46                     & 22.78                    & 8.31                     & 11.89                    \\
PGD-AT                                & 8/255                & 57.43                    & 53.71                     & 37.66                    & 24.81                    & 24.33                    & 50.90                     & 30.07                    & 13.51                    & 19.62                    & 47.43                     & 23.18                    & 7.40                     & 11.64                    \\
ATES                                  & 8/255                & 57.54                    & 53.62                     & 37.05                    & 25.08                    & 24.72                    & 50.84                     & 29.18                    & 13.75                    & 19.42                    & 47.35                     & 22.89                    & 7.59                     & 11.40                    \\
ExAT-PGD                              & 9/255                & 57.46                    & 53.56                     & 38.48                    & 25.25                    & 24.93                    & 51.43                     & 30.60                    & 15.12                    & 20.40                    & 48.15                     & 24.21                    & 8.37                     & 12.47                    \\
ExAT-AWP        & 10/255                                                             & 57.76          & 53.46                                                                & 37.84                                                              & 25.55                                                                & 25.27                                                          & 50.42                                                                 & 30.39                                                               & 14.98                                                                 & 19.72                                                               & 46.99                                                                 & 24.48                                                               & 9.07                                                                  & 12.68                                     \\                         

AWP                                   & 8/255                & 58.81                    & 54.13                     & 37.92                    & 25.51                    & 25.30                    & 50.72                     & 30.40                    & 14.71                    & 19.82                    & 46.66                     & 23.96                    & 8.68                     & 12.44                    \\
AWP (with AutoAug.)                   & 8/255                                                              & 59.88                    & 55.62                                                                & 39.10                                                               & 25.81                                                                & 25.52                                                          & 52.75                                                                 & 31.11                                                               & 14.80                                                                 & 20.24                                                               & 49.44                                                                 & 24.99                                                               & 8.72                                                                  & 12.80                                                               \\

\textbf{Ours (with AutoAug)} & 16/255               & \textbf{60.27}           & \textbf{56.27}                     & \textbf{40.24}           & \textbf{26.41}           & \textbf{26.00}           & \textbf{53.86}                     & \textbf{33.78}           & \textbf{16.28}           & \textbf{21.47}           & \textbf{51.11}                     & \textbf{28.02}           & \textbf{10.47}           & \textbf{14.60}           \\

\textcolor{ao}{$^{\textrm{Gain w.r.t. AWP (with AutoAug)}}$}  & & \textcolor{ao}{$^{~~+0.39}$} & \textcolor{ao}{$^{~~+0.65}$} & \textcolor{ao}{$^{~~+1.14}$} & \textcolor{ao}{$^{~~+0.60}$} & \textcolor{ao}{$^{~~+0.48}$} & \textcolor{ao}{$^{~~+1.11}$} & \textcolor{ao}{$^{~~+2.67}$} & \textcolor{ao}{$^{~~+1.48}$} & \textcolor{ao}{$^{~~+1.23}$} &\textcolor{ao}{$^{~~+1.67}$} & \textcolor{ao}{$^{~~+3.03}$} & \textcolor{ao}{$^{~~+1.75}$} & \textcolor{ao}{$^{~~+1.80}$} \\
\bottomrule
\end{tabular}}
% \vspace{-0.3cm}
\end{table*}
\begin{table}[b]
\vspace{-0.5cm}
\caption{Prediction accuracy (\%) of PreActResNet-18 models trained using TRADES-AWP and OA-AT on $\ell_2$ adversaries. }
%\vspace{-0.3cm}
\setlength\tabcolsep{2pt}

\label{table:l2_PRN18_mini}
\begin{tabular}{l|c|c|c|c|c}
                                
                                   \toprule
{\textbf{Dataset}} & {\textbf{Method}} & \textbf{Clean}       & \textbf{AA@0.5}       & \textbf{Square@0.75}      & \textbf{Square@1}      \\
\midrule
                                                           & TRADES-AWP                                                & 88.45       & 71.34        & 71.19            & 64.21         \\
{\textbf{CIFAR-10}}                                 & OA-AT (Ours)                                               & \textbf{89.13}       & \textbf{71.40}         & \textbf{73.33}            & \textbf{66.48}         \\
\midrule
                                                           & TRADES-AWP                                                & 70.38       & 41.96        & 46.77            & 38.62         \\
{\textbf{CIFAR-100}}                                 & OA-AT (Ours)                                               & \textbf{70.41}       & \textbf{44.70}         & \textbf{49.34}            & \textbf{41.58}   \\

\bottomrule

\end{tabular}
\vspace{-0.3cm}
\end{table}
\subsection{$\ell_2$ Threat Model} OA-AT indeed works well when trained on $\ell_2$ adversaries, as shown in Table-\ref{table:l2_PRN18_mini}. While the standard perturbation bound considered for $\ell_2$ norm is 0.5, we show significant improvements for $\varepsilon$= 0.75, 1 as well. We obtain consistent gains at large $\varepsilon$ bounds while achieving similar clean accuracy and robust accuracy at the lower bound of 0.5. On CIFAR-100, we obtain 2.5\%-3\% gains across perturbation bounds of 0.5, 0.75 and 1.

\subsection{AWP+ results}
\label{AWP_plus_suppl}
Since the proposed method utilizes additional techniques to overcome overfitting and improve generalization, we generate improved baselines as well, using the same techniques, in order to facilitate  a fair comparison. In Table-\ref{table:awp_plus}, we present results of improved AWP \cite{wu2020adversarial} baselines by using AutoAugment \cite{cubuk2018autoaugment,stutz2021relating}, Weight Averaging \cite{izmailov2018averaging} and Label Smoothing + Warmup. As previously seen from evaluations as reported in Table-\ref{table:cifar10_main}, we again observe that the proposed method OA-AT consistently outperforms all the AWP+ baselines presented in Table-\ref{table:awp_plus}. Further, we present OA-AT and AWP \cite{wu2020adversarial} with and without Weight Averaging \cite{izmailov2018averaging} in Table-\ref{table:SWA_table}, by training WideResNet-34-10 models. We observe that Weight Averaging does not lead to significant gains for both OA-AT as well as AWP \cite{wu2020adversarial}.

\begin{table}[h]
\centering
\caption{\textbf{Effect of Weight Averaging \cite{izmailov2018averaging} on AWP \cite{wu2020adversarial} and OA-AT}: Performance ($\%$) of WideResNet-34-10 models trained using the proposed defense OA-AT and AWP \cite{wu2020adversarial}, against GAMA-PGD100 \cite{sriramanan2020gama}  and  Square \cite{andriushchenko2019square} attacks with and without using Weight Averaging \cite{izmailov2018averaging}.}

\setlength\tabcolsep{3pt}

%\resizebox{\linewidth}{!}{
\label{table:SWA_table}
\begin{tabular}{l|ccc|ccc}

% \vspace{-0.3cm}
\toprule
                & \multicolumn{3}{c}{\textbf{CIFAR-10, WRN-34-10}}                                                                                                                                                                           & \multicolumn{3}{|c}{\textbf{CIFAR-100, WRN-34-10}}                                                                                               \\
                \cline{2-7}
                
\small{\textbf{Method}} & \small{\textbf{\begin{tabular}[c]{@{}c@{}}Clean \\ Acc\end{tabular}}} &  \small{\textbf{\begin{tabular}[c]{@{}c@{}}GAMA \\ (8/255)\end{tabular}}} & \small{\textbf{\begin{tabular}[c]{@{}c@{}}Square \\ (16/255)\end{tabular}}}  & \small{\textbf{\begin{tabular}[c]{@{}c@{}}Clean \\ Acc\end{tabular}}} &  \textbf{\textbf{\begin{tabular}[c]{@{}c@{}}GAMA \\ (8/255)\end{tabular}}} & \small{\textbf{\begin{tabular}[c]{@{}c@{}}Square \\ (16/255)\end{tabular}}} \\
\midrule
AWP (without WA)    & 85.36          & 56.34                 & 31.70                    & 62.78          & 29.82                 & 15.70                    \\
AWP+ (with WA)      & 85.52          & 56.42                 & 32.41                    & 62.73          & 29.92                 & 15.85                    \\
OAAT (without WA)   & 85.28          & 58.19                 & 36.75                    & 65.53          & 30.59                 & 18.06                    \\
Ours (OAAT with WA) & 85.32          & 58.48                 & 36.93                    & 65.73          & 30.90                 & 18.47   \\
\bottomrule
\end{tabular} %}
% \end{minipage}
\hfill
\vspace{-0.3cm}
\end{table}
\begin{table}
\centering
\caption{\textbf{Improvements to the AWP baseline:}  Performance (\%) of models trained by applying AutoAugment \cite{cubuk2018autoaugment}, Label Smoothing + Warmup and Weight Averaging \cite{izmailov2018averaging} to the AWP baseline \cite{wu2020adversarial}, against GAMA-PGD100 \cite{sriramanan2020gama}  and  Square \cite{andriushchenko2019square} attacks. Results on the CIFAR-10, CIFAR-100 and SVHN datasets are reported using different $\varepsilon$ bounds. }

\setlength\tabcolsep{1pt}
\label{table:awp_plus}
\resizebox{\linewidth}{!}{
\begin{tabular}{ccc|ccc|c}
\toprule
\multirow{3}{*}{\textbf{\begin{tabular}[c]{@{}c@{}}Auto-\\Augment \\ probability\end{tabular}}} & \multirow{3}{*}{\textbf{\begin{tabular}[c]{@{}c@{}}Label \\Smoothing \\ + Warmup\end{tabular}}} & \multirow{3}{*}{\textbf{\begin{tabular}[c]{@{}c@{}}Weight \\ Averaging\end{tabular}}} & \multicolumn{3}{c|}{\textbf{Metrics of interest}}                                                                                                      & \textbf{Others}                                                  \\
\cmidrule{4-7} 
                                                                                             &                                                                                               &                                                                                       & \textbf{~~~~Clean~~~} & \textbf{\begin{tabular}[c]{@{}c@{}}~~~GAMA~~~\\ (8/255)\end{tabular}} & \textbf{\begin{tabular}[c]{@{}c@{}}~~~Square~~~\\ (16/255)\end{tabular}} & \textbf{\begin{tabular}[c]{@{}c@{}}~~~GAMA~~~\\ (16/255)\end{tabular}} \\

                                                                    \midrule
\multicolumn{7}{c}{\textbf{CIFAR-10 (WRN-34-10), 200 epochs}}      \\
\midrule
0                                                                                         & $\times$                                                                                            & $\times$                                                                                    & 85.36          &  56.34                                                      &  31.54                                                           &   23.74                                                       \\
0                                                                                         & $\times$                                                                                            & \checkmark                                                                                   & 85.52          &  56.42                                                         &   32.41                                                            &  24.04                                                          \\

1                                                                                         & $\times$                                                                                            & $\times$                                                                                    & 87.36          & 52.62                                                       & 29.83                                                              & 19.39                                                            \\
1                                                                                         & $\times$                                                                                            & \checkmark                                                                                   & 86.75       & 53.62                                                           &    30.11                                                           & 20.41                                                            \\
\midrule
\multicolumn{7}{c}{\textbf{CIFAR-100 (ResNet-18), 110 epochs}}     \\
\midrule
0                                                                                         & $\times$                                                                                            & $\times$                                                                                    & 58.81          & 25.51                                                           & 12.44                                                              & 8.68                                                             \\
0.5                                                                                         & $\times$                                                                                            & $\times$                                                                                    & 59.88          & 25.81                                                           & 12.80                                                              & 8.72                                                             \\
0                                                                                         & \checkmark                                                                                           & \checkmark                                                                                   & 58.99          & 26.07                                                           & 13.10                                                              & 8.98                                                             \\
0.5                                                                                         & \checkmark                                                                                           & $\times$                                                                                    & 59.82          & 25.39                                                           & 13.04                                                              & 8.62                                                             \\
\midrule
\multicolumn{7}{c}{\textbf{CIFAR-100 (PreActResNet-18), 200 epochs}}    \\
\midrule
0                                                                                         & $\times$                                                                                            & $\times$                                                                                    & 58.85          & 25.58                                                           & 12.39                                                              & 9.01                                                             \\
0.5                                                                                         & \checkmark                                                                                           & $\times$                                                                                    & 62.10          & 25.99                                                           & 13.27                                                              & 8.91                                                             \\
0.5                                                                                         & \checkmark                                                                                           & \checkmark                                                                                   & 62.11          & 26.21                                                           & 13.26                                                              & 9.21                                                             \\
0                                                                                         & \checkmark                                                                                           & $\times$                                                                                    & 59.70          & 26.61                                                           & 13.80                                                              & 9.70                                                             \\
0                                                                                         & \checkmark                                                                                           & \checkmark                                                                                   & 59.97          & 26.90                                                           & 13.74                                                              & 9.95                                                             \\
\midrule
\multicolumn{7}{c}{\textbf{CIFAR-100 (WRN-34-10), 110 epochs}}        \\
\midrule
0                                                                                         & $\times$                                                                                            & $\times$                                                                                    & 62.41          & 28.98                                                           & 14.68                                                              & 10.98                                                            \\
0                                                                                         & $\times$                                                                                            & \checkmark                                                                                   & 61.72          & 29.78                                                           & 15.32                                                              & 11.15                                                            \\
0.5                                                                                         & $\times$                                                                                            & $\times$                                                                                    & 61.33          & 29.22                                                           & 15.18                                                              & 10.94                                                            \\
0                                                                                         & \checkmark                                                                                           & $\times$                                                                                    & 62.78          & 29.82                                                           & 15.70                                                              & 11.45                                                            \\
0                                                                                        & \checkmark                                                                                           & \checkmark                                                                                   & 62.73          & 29.92                                                           & 15.85                                                              & 11.55                                                            \\
0.5                                                                                         & \checkmark                                                                                           & \checkmark                                                                                   & 62.23          & 29.36                                                           & 15.47                                                              & 11.20                                                            \\
\midrule
\multicolumn{7}{c}{\textbf{SVHN (PreActResNet-18), 110 epochs}}   \\
\midrule
0                                                                                         & $\times$                                                                                            & $\times$                                                                                    & 91.91          & 75.92                                                           & 35.78                                                              & 30.70                                                            \\
0.5                                                                                         & $\times$                                                                                            & $\times$                                                                                    & 90.99          & 75.37                                                           & 36.42                                                              & 31.02                                                            \\
0.5                                                                                         & $\times$                                                                                            & \checkmark                                                                                   & 92.21          & 72.31                                                           & 36.02                                                              & 30.80                                                            \\
1                                                                                         & $\times$                                                                                            & $\times$                                                                                    & 89.97          & 75.08                                                           & 38.47                                                              & 31.34                                                            \\
1                                                                                         & $\times$                                                                                            & \checkmark                                                                                   & 89.71          & 74.73                                                           & 38.41                                                              & 31.15          \\
\bottomrule
\end{tabular}}
\end{table}

\section{Gradient Masking Checks}
\label{sec:grad_mask_checks}

\begin{figure*}[!h]
\centering
        \includegraphics[width=\linewidth]{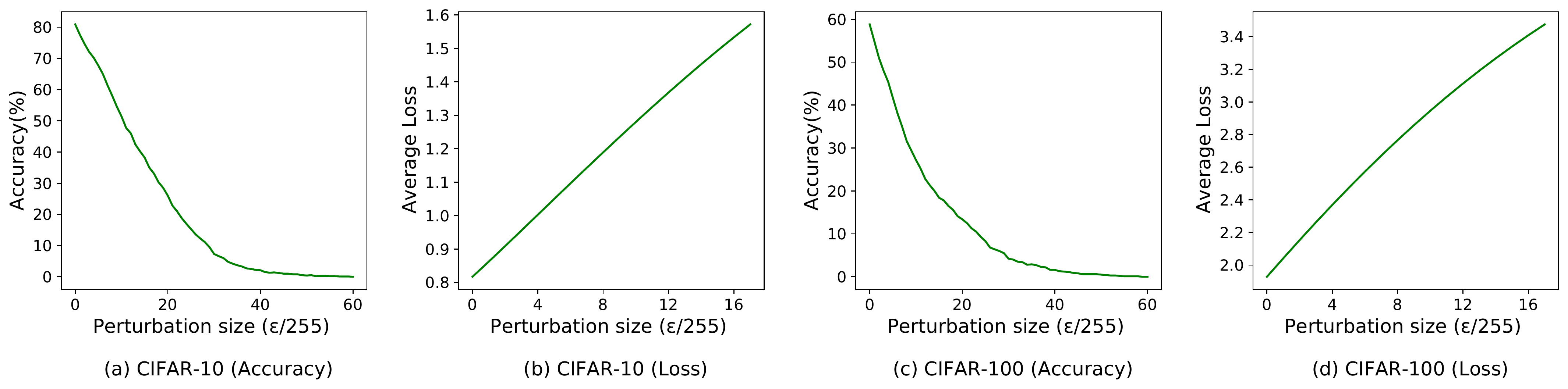}
        %\vspace{-0.3cm} 
        \caption{Accuracy and Loss plots on a $1000$-sample class-balanced subset of the respective test-sets of CIFAR-10 and CIFAR-100 datasets.
        (a, c) Plots showing the trend of Accuracy ($\%$) against PGD-$7$ step attacks across variation in attack perturbation bound ($\varepsilon$) on CIFAR-10 and CIFAR-100 datasets with ResNet-18 architecture. As the perturbation bound increases, accuracy against white-box attacks goes to $0$, indicating the absence of gradient masking \cite{athalye2018obfuscated} (b, d) Plots showing the variation of Cross-Entropy Loss on FGSM attack \cite{goodfellow2014explaining} against variation in the attack perturbation bound ($\varepsilon$). As the perturbation bound increases, loss increases linearly, indicating the absence of gradient masking \cite{athalye2018obfuscated}}
        \label{fig:loss_accuracy}
        % \vspace{-0.5cm}
\end{figure*}

As discussed by Athalye et al. \cite{athalye2018obfuscated}, we present various checks to ensure the absence of Gradient Masking in the proposed defense. In Fig.\ref{fig:loss_accuracy} (a,c), we observe that the accuracy of the proposed defense on the CIFAR-10 and CIFAR-100 datasets monotonically decreases to zero against 7-step PGD white-box attacks as the perturbation budget is increased. This shows that gradient based attacks indeed serve as a good indicator of robust performance, as strong adversaries of large perturbation sizes achieve zero accuracy, indicating the absence of gradient masking. In Fig.\ref{fig:loss_accuracy} (b,d), we plot the Cross-Entropy loss against FGSM attacks with varying perturbation budget. We observe that the loss increases linearly, thereby suggesting that the first-order Taylor approximation to the loss surface indeed remains effective in the local neighbourhood of sample images, again indicating the absence of gradient masking.

We verify that the model achieves higher robust accuracy against weaker Black-box attacks, when compared to strong gradient based attacks such as GAMA or AutoAttack in Tables-\ref{table:cifar10_full}, \ref{table:cifar100_full}. We also observe that adversaries that conform to larger constraint sets are stronger than their counterparts that are restricted to smaller epsilon bounds, as expected.
\vspace{0.3cm}
%\vspace{-0.3cm}

In Table-\ref{table:robust_eval}, we perform exhaustive evaluations using various attack techniques to further verify the absence of gradient masking. In addition to AutoAttack \cite{croce2020reliable} which in itself consists of an ensemble of four attacks (AutoPGD with Cross-Entropy and Difference-of-Logits loss, the FAB attack \cite{croce2019minimally} and Square Attack \cite{andriushchenko2019square}), we present evaluations against strong multi-targeted attacks such as GAMA-MT \cite{sriramanan2020gama} and the MDMT attack \cite{jiang2020imbalanced} which specifically target other classes during optimization. We also consider the untargeted versions of the latter two attacks, the GAMA-PGD and MD attack respectively. We also present robustness against the ODS attack \cite{tashiro2020diversity} with 100 restarts, which diversifies the input random noise based on the output predictions in order to obtain results which are less dependent on the sampled random noise used for attack initialization. Next, the Logit-Scaling attack \cite{carlini2016defensive,hitaj2021evaluating} helps yield robust evaluations that are less dependent on the exact scale of output logits predicted by the network, and is seen to be effective on some defenses which exhibit gradient masking. However, we observe that the proposed method is robust against all such attacks, with the lowest accuracy being attained on the AutoAttack ensemble. 
\begin{table}
\begin{center}
\caption{\textbf{Evaluation against various attacks constrained within a perturbation bound of $\varepsilon=8/255$ on CIFAR-10:} Performance ($\%$) of the proposed defense OA-AT on ResNet-18 architecture against various attacks (sorted by Robust Accuracy) to ensure the absence of gradient masking. \\ $^\dagger$Includes 5000-queries of Square attack.}
\setlength\tabcolsep{3pt}
\resizebox{0.95\linewidth}{!}{
\label{table:robust_eval}
\begin{tabular}{l|cc|c}
\toprule
\textbf{Attack}                    & \textbf{No. of Steps} & \textbf{No. of restarts} & \textbf{Robust Accuracy ($\%$)}  \\
\midrule
AutoAttack$^{\dagger}$   \cite{croce2020reliable}                         & 100     & 20                          & \textbf{50.88} \\
GAMA-MT  \cite{sriramanan2020gama}                          & 100                      & 5                           & 50.90          \\
ODS (98 +2 steps)  \cite{tashiro2020diversity}        & 100                      & 100                         & 50.94          \\
MDMT attack  \cite{jiang2020imbalanced}                      & 100                      & 10                           & 51.19          \\
Logit-Scaling attack \cite{carlini2016defensive,hitaj2021evaluating} & 100                      & 20                          & 51.26          \\
GAMA-PGD   \cite{sriramanan2020gama}                        & 100                      & 1                           & 51.40          \\
MD attack \cite{jiang2020imbalanced}                         & 100                      & 1                           & 51.47          \\
PGD-50 (1000 RR) \cite{madry-iclr-2018}                  & 50                       & 1000                        & 55.37          \\
PGD-1000 \cite{madry-iclr-2018}                           & 1000                     & 1                           & 56.15          \\ 
\bottomrule
\end{tabular}}
\end{center}
%\vspace{-0.2cm}
\end{table}
\begin{table}[!h]
\vspace{-0.4cm}
\caption{\textbf{Evaluation against an ensemble of AutoAttack (AA) \cite{croce2020reliable} and Multi-Targeted (MT) attack \cite{gowal2020uncovering}:} Evaluating against the Multi-Targeted (MT) attack along with AutoAttack \cite{croce2020reliable} leads to a marginal decrease in robust accuracy, thus showing that AutoAttack \cite{croce2020reliable} is sufficient to obtain a reliable estimate of robustness.}
\label{table:AA_plus_MT}

\begin{center}

\resizebox{0.8\linewidth}{!}{
\begin{tabular}{c|ccccc}
\toprule
\multirow{2}{*}{\textbf{\begin{tabular}[c]{@{}c@{}}Method\end{tabular}}} & \multirow{2}{*}{\textbf{\begin{tabular}[c]{@{}c@{}}~~Clean~~ \\ Accuracy\end{tabular}}} & \multirow{2}{*}{\textbf{\begin{tabular}[c]{@{}c@{}}~~AA~~ \\ (8/255)\end{tabular}}} & \multirow{2}{*}{\textbf{\begin{tabular}[c]{@{}c@{}}~~MT ~~\\ (8/255)\end{tabular}}} & \multirow{2}{*}{\textbf{\begin{tabular}[c]{@{}c@{}}~~AA + MT~~\\(8/255)\end{tabular}}} & \multirow{2}{*}{\textbf{\begin{tabular}[c]{@{}c@{}}~~SQ + RS~~ \\ (16/255)\end{tabular}}} \\ 
& & & & & \\
\midrule

\multicolumn{1}{l}{} & \multicolumn{5}{c}{\textbf{CIFAR-10, WRN-34-10}}                                                                    \\
\midrule
AWP                  & \textbf{85.36} & 56.17               & 56.17               & 56.15                    & 30.87                                               \\
Ours                 & 85.32          & \textbf{58.04}      & \textbf{58.06}      & \textbf{58.03}           & \textbf{35.31}                                      \\

\midrule
\multicolumn{1}{l}{} & \multicolumn{5}{c}{\textbf{CIFAR-100, WRN-34-10}}                                                                    \\
\midrule
AWP                  & 62.73          & 29.92               & 29.92               & 29.91                    & 14.96         \\
Ours                 & \textbf{65.73} & \textbf{30.35}      & \textbf{30.49}      & \textbf{30.34}           & \textbf{17.15}                                      \\

\bottomrule
\end{tabular}}
\end{center}
\vspace{-0.4cm}
\end{table}

Further, we evaluate the model on PGD 50-step attack run with 1000 restarts. The robust accuracy saturates with increasing restarts, with the final accuracy still being higher than that achieved on AutoAttack. Lastly, we observe that the PGD-1000 attack is not very strong, confirming that the accuracy does not continually decrease as the number of steps used in the attack increases. Thus, we observe that the proposed approach is robust against a diverse set of attack methods, thereby confirming the absence of gradient masking and verifying that the model is truly robust. 

We also evaluate the WideResNet-34-10 model trained using OA-AT (proposed approach) and AWP \cite{wu2020adversarial} on CIFAR-10 and CIFAR-100 datasets, against an ensemble of AutoAttack \cite{croce2020reliable} and Multi-Targeted attacks \cite{gowal2020uncovering} in Table-\ref{table:AA_plus_MT}. We observe that using Multi-Targeted attack along with AutoAttack only leads to a drop of 0.01-0.02 $\%$ in the robust accuracy, suggesting that AutoAttack \cite{croce2020reliable} is sufficient to obtain a reliable estimate of robustness. 

\section{Details on Contrast Calculation}
\begin{figure*}[t]
\begin{minipage}{0.48\linewidth}
\centering
        \includegraphics[width=\linewidth]{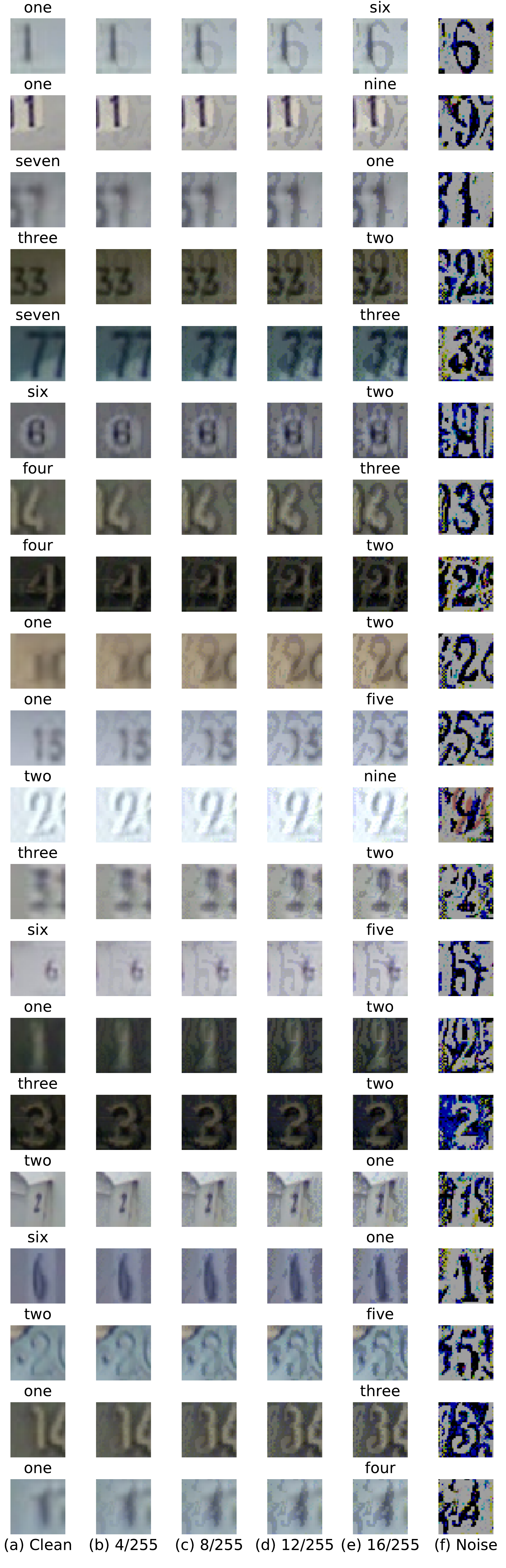}
        \vspace{-0.2cm}
        \caption{\textbf{SVHN, Low-Contrast}}
        \label{fig:svhn_LC}
        \end{minipage}
        \hfill
\begin{minipage}{0.48\linewidth}
\centering
        \includegraphics[width=\linewidth]{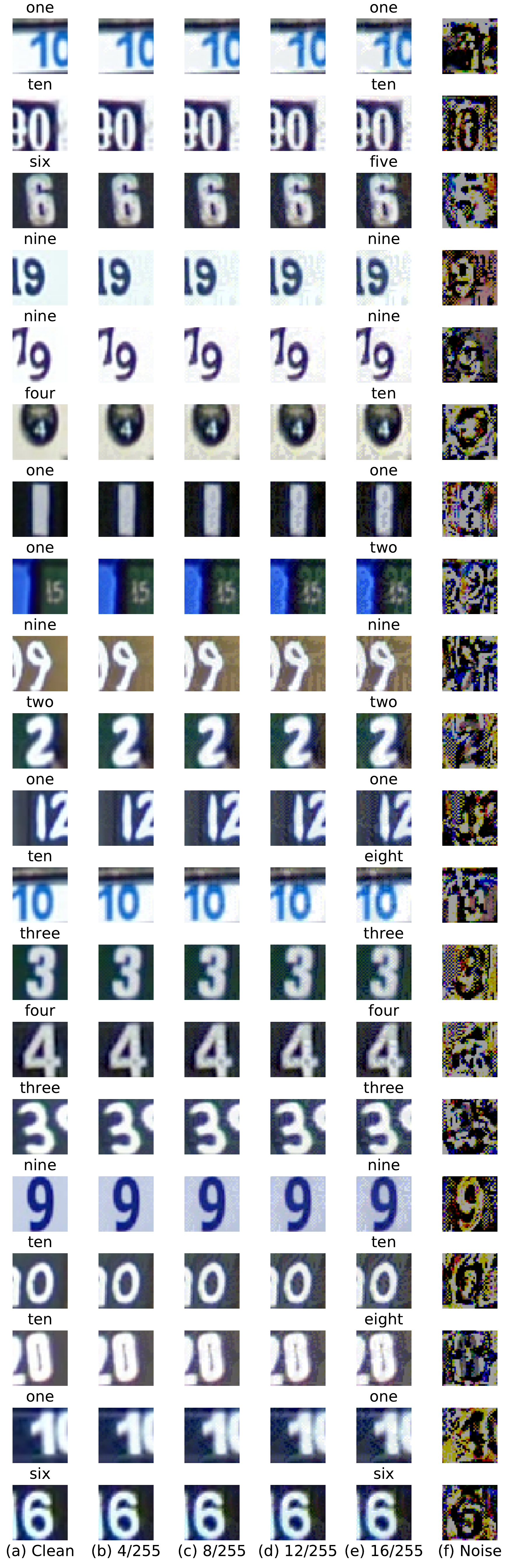}
        \vspace{-0.1cm}
        \caption{\textbf{SVHN, High Contrast}}
        \label{fig:svh_HC}
 \end{minipage}
\end{figure*}

\begin{figure*}[t]
\begin{minipage}{0.48\linewidth}
\centering
        \includegraphics[width=\linewidth]{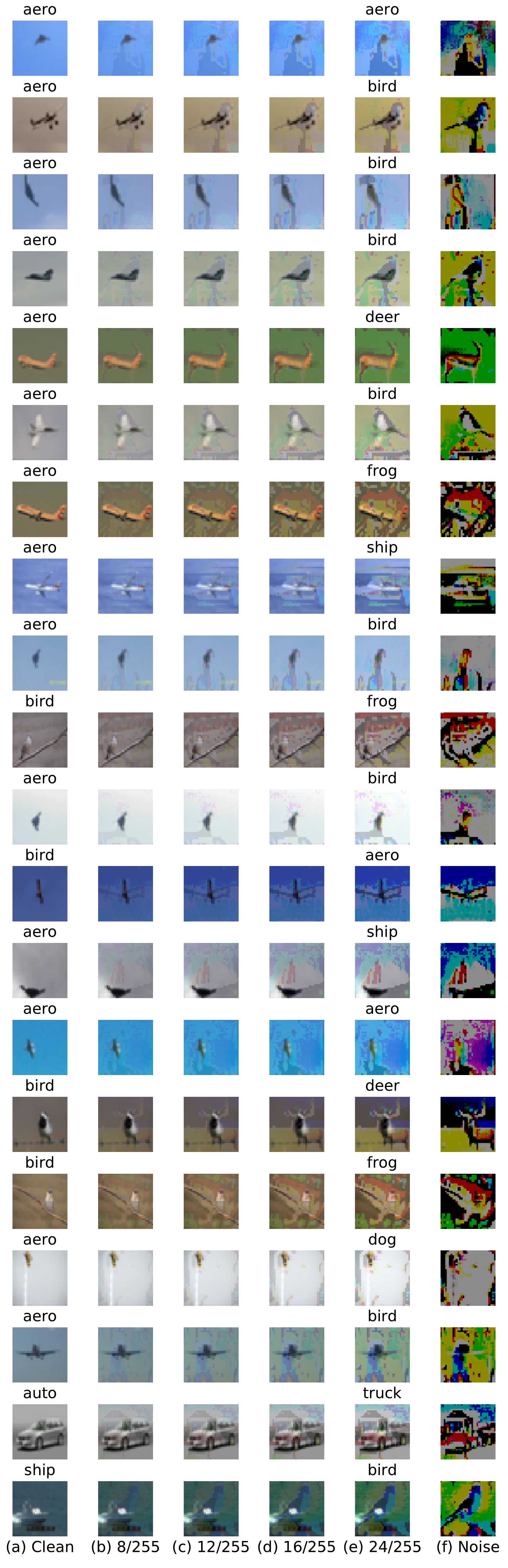}
         \vspace{-0.0cm}
        \caption{\textbf{CIFAR-10 Low Contrast}}
        \label{fig:CIFAR10_LC}
        \end{minipage}
        \hfill
\begin{minipage}{0.48\linewidth}
\centering
        \includegraphics[width=\linewidth]{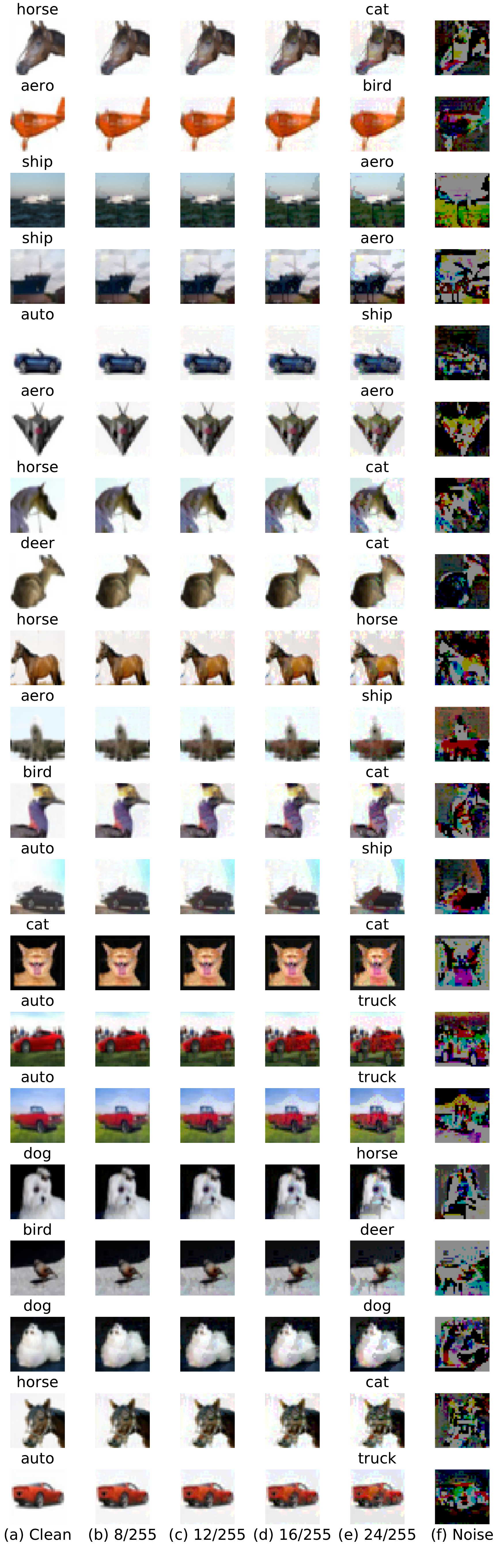}
         \vspace{-0.1cm}
        \caption{\textbf{CIFAR-10 High Contrast}}
        \label{fig:CIFAR10_HC}
\end{minipage}
\end{figure*}

\begin{figure*}[t]
\begin{minipage}{0.48\linewidth}
\centering
        \includegraphics[width=\linewidth]{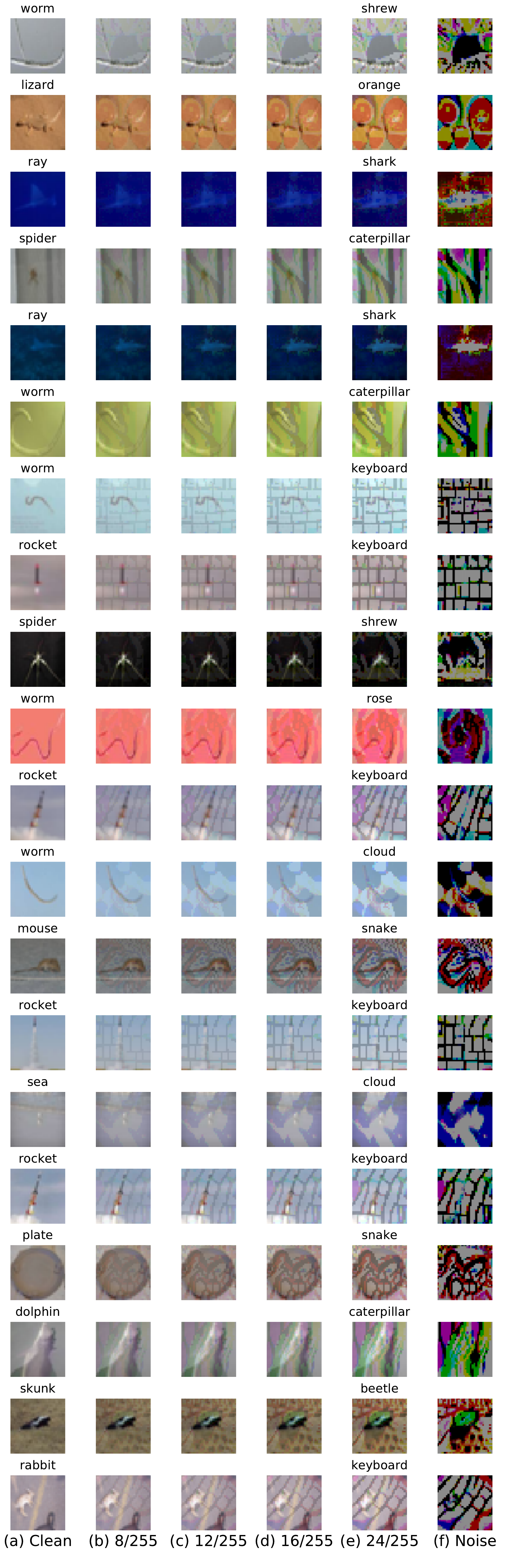}
         \vspace{-0.1cm}
        \caption{\textbf{CIFAR100 Low Contrast}}
        \label{fig:CIFAR100_LC}
        \end{minipage}
        \hfill
\begin{minipage}{0.48\linewidth}
\centering
        \includegraphics[width=\linewidth]{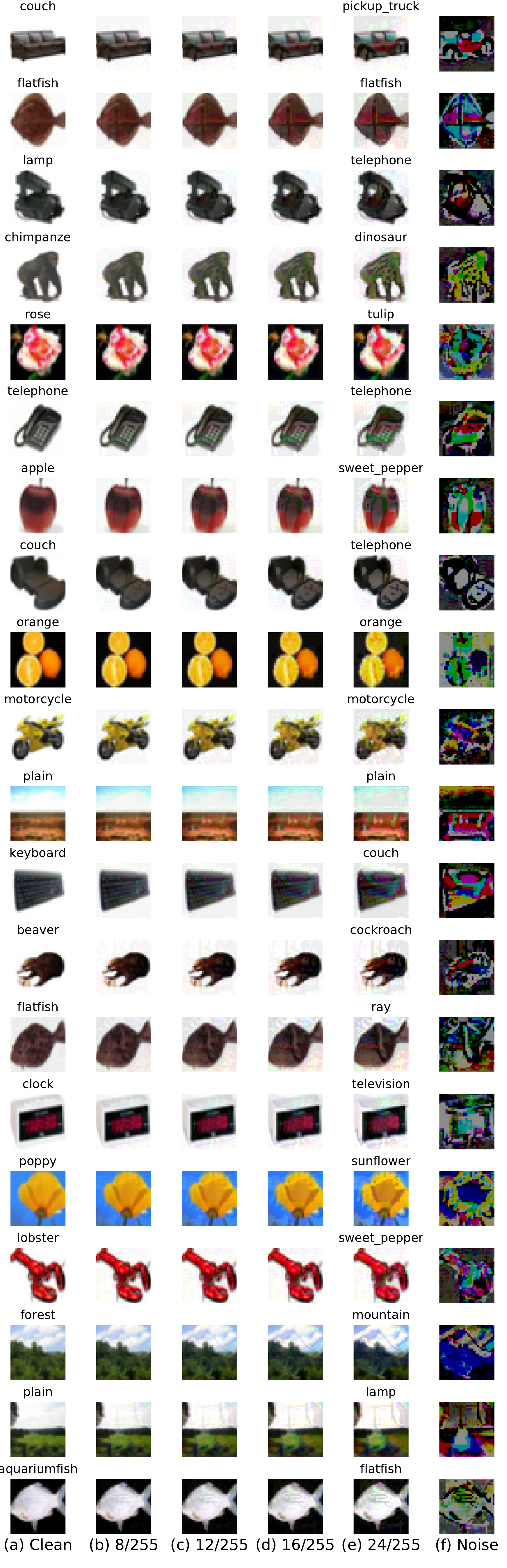}
         \vspace{-0.1cm}
        \caption{\textbf{CIFAR100 High Contrast}}
        \label{fig:CIFAR100_HC}
\end{minipage}
\end{figure*}

In order to determine the contrast level for a given image, the mean absolute deviation of each pixel is first computed for the three RGB color channels independently. Following this, top 20\% of pixels which correspond to the highest mean absolute deviations averaged over the three channels are selected. The variance in intensities over these selected pixels, averaged over the three channels, is used as a measure of contrast for the image. We sort images in order of increasing contrast and split the dataset into 10 bins for the evaluations in Fig.\ref{fig:subset_eval}. We present the Low and High Contrast images on SVHN, CIFAR-10 and CIFAR-100 datasets respectively in Fig.\ref{fig:svhn_LC}, \ref{fig:svh_HC}, \ref{fig:CIFAR10_LC}, \ref{fig:CIFAR10_HC}, \ref{fig:CIFAR100_LC} and \ref{fig:CIFAR100_HC}.

\section{Sensitivity towards Hyperparameters}

\begin{figure}[!h]
\centering
        \includegraphics[width=\linewidth]{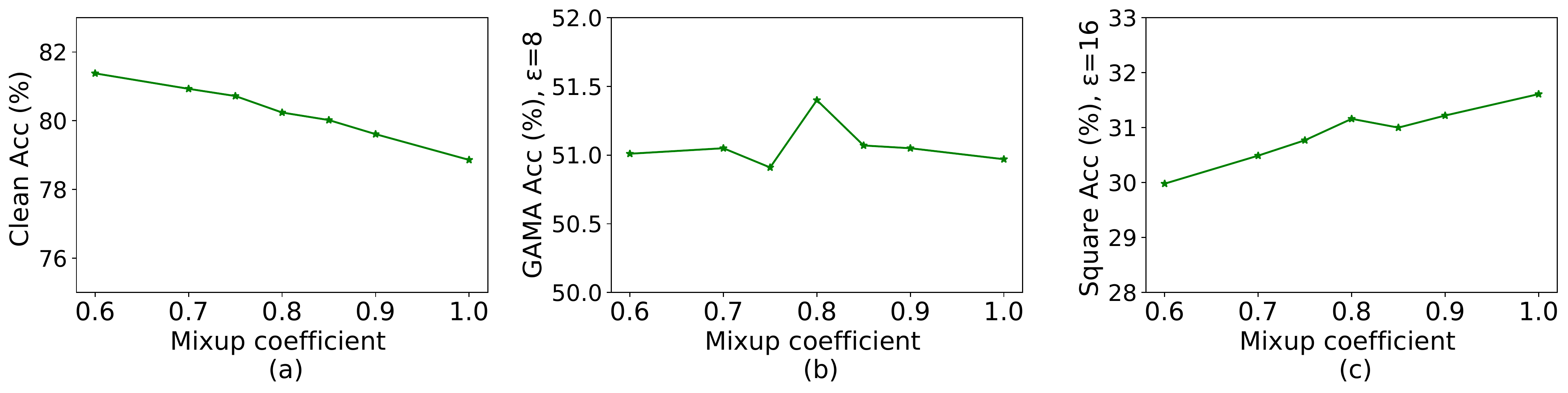}
        \caption{\textbf{Sensitivity of the proposed approach against variation in Mixup coefficient:} (a) Clean Accuracy ($\%$), (b) Accuracy ($\%$) against GAMA-PGD 100-step attack \cite{sriramanan2020gama} at $\varepsilon=8/255$ and (c) Accuracy ($\%$) against Square Attack \cite{andriushchenko2019square} at $\varepsilon=16/255$ are reported on the CIFAR-10 dataset for ResNet-18 architecture. The optimal setting chosen is mixup coefficient of 0.8.}
        \label{fig:sens_mixup_coeff}

\centering
        \includegraphics[width=\linewidth]{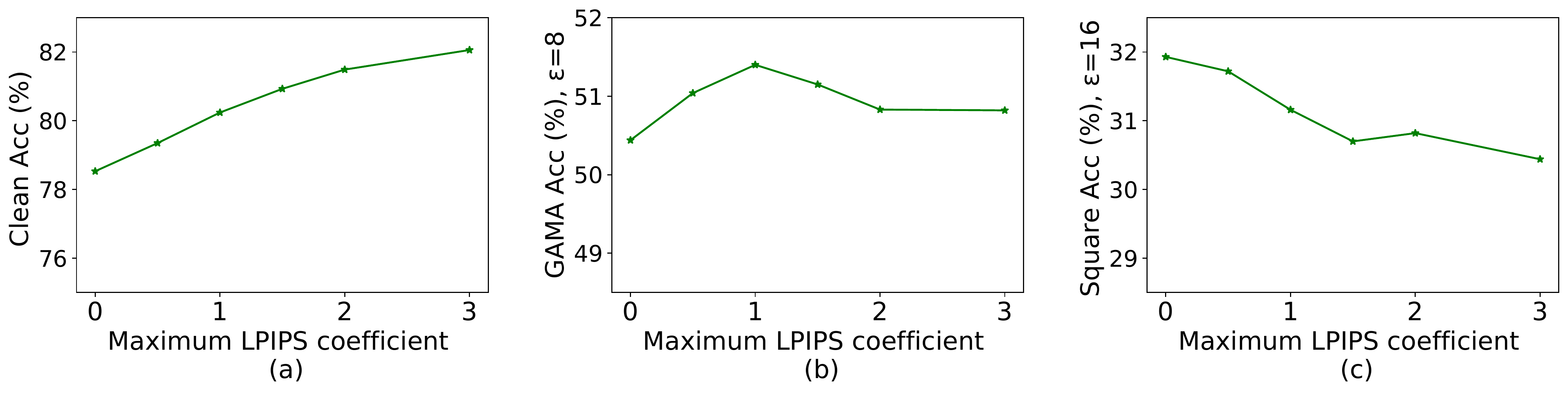}
        \caption{\textbf{Sensitivity of the proposed approach against variation in Maximum LPIPS coefficient:} (a) Clean Accuracy ($\%$), (b) Accuracy ($\%$) against GAMA-PGD 100-step attack \cite{sriramanan2020gama} at $\varepsilon=8/255$ and (c) Accuracy ($\%$) against Square Attack \cite{andriushchenko2019square} at $\varepsilon=16/255$ are reported on the CIFAR-10 dataset for ResNet-18 architecture. The optimal setting chosen is maximum LPIPS coefficient of 1.}
        \label{fig:sens_lpips_coeff}

\centering
        \includegraphics[width=\linewidth]{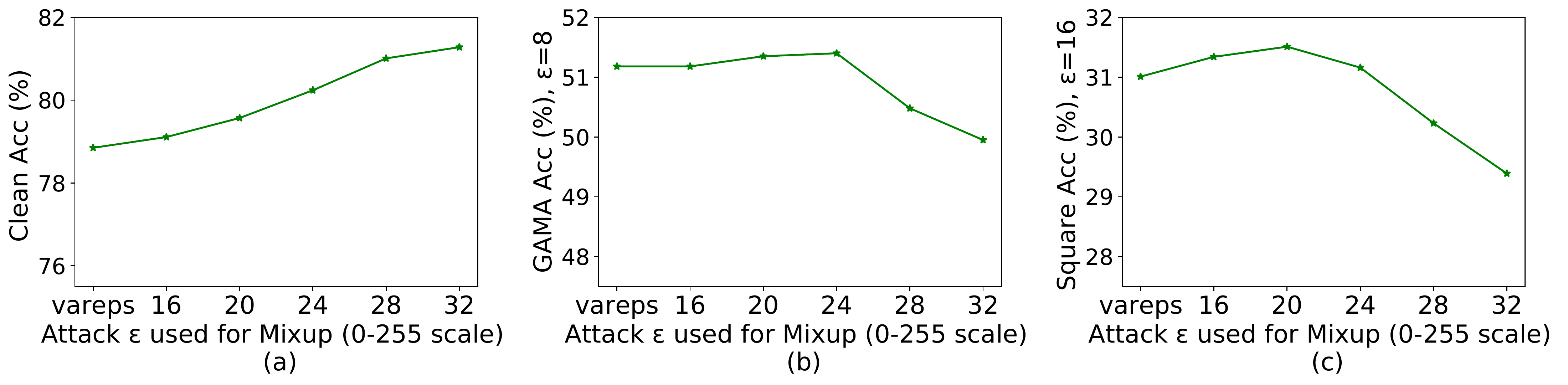}
        \caption{\textbf{Sensitivity of the proposed approach against variation in $\varepsilon$ used in mixup iteration:} (a) Clean Accuracy ($\%$), (b) Accuracy ($\%$) against GAMA-PGD 100-step attack \cite{sriramanan2020gama} at $\varepsilon=8/255$ and (c) Accuracy ($\%$) against Square Attack \cite{andriushchenko2019square} at $\varepsilon=16/255$ are reported on the CIFAR-10 dataset for ResNet-18 architecture. The optimal setting chosen is $\varepsilon=24$ for mixup.}
        \label{fig:sens_mixup_eps}
\end{figure}

We check the sensitivity of the proposed method across variation in different hyperparameters on the CIFAR-10 dataset with ResNet-18 model architecture using a 110 epoch training schedule. The value of the mixup coefficient is varied from 0.6 to 1 as seen in Fig.\ref{fig:sens_mixup_coeff}. On increasing the value of mixup coefficient, clean accuracy drops due to the presence of Oracle-Sensitive adversarial examples. While a lower value of mixup coefficient helps in improving clean accuracy, it makes the attack weaker, resulting in a lower robust accuracy. We visualize the effect of changing the maximum value of LPIPS coefficient in Fig.\ref{fig:sens_lpips_coeff}. Using a higher LPIPS coefficient helps in boosting the clean accuracy while dropping the adversarial accuracy, while a low value close to zero drops both clean as well as robust accuracy due to the presence of oracle sensitive examples. Finally, we show the effect of changing the $\varepsilon$ (referred to as $\varepsilon_{ref}$) used in the mixup iteration. We find that using a higher value of $\varepsilon_{ref}$ in mixup iteration leads to weak attack since we project every perturbation to a much lower epsilon value while training, resulting in a higher clean accuracy and lower robust accuracy. However, a higher value of $\varepsilon_{ref}$ also leads to a more reliable estimate of the Oracle prediction, thereby leading to improved robust accuracy at intermediate values of $\varepsilon_{ref}$. Overall, we observe that OA-AT is less sensitive to hyperparameter changes.

\begin{table}[t]
\begin{center}
\caption{\textbf{Transferability of hyperparameters across datasets:} Performance (\%) of the proposed method using different sets of training hyperparameters when compared to the AWP \cite{wu2020adversarial} baseline. The setting, \textit{tuned} indicates that the hyperparameters have been specifically tuned for the given dataset, while \textit{no tuning} indicates that we use the same set of hyperparameters that were found on a different dataset (CIFAR-10). The performance of the \textit{tuned} case is only marginally better than the \textit{no tuned} case indicating that the proposed method is not sensitive to changes in hyperparameters.}
\vspace{0.1cm}
%\resizebox{1.0\linewidth}{!}{
\label{table:datasets_transfer}
\begin{tabular}{lcccc}
\toprule
\multicolumn{1}{l}{\textbf{Method}}            & \textbf{Clean}    & \multirow{2}{*}{\textbf{\begin{tabular}[c]{@{}c@{}}~~GAMA~~ \\ (4/255)\end{tabular}}}    & \multirow{2}{*}{\textbf{\begin{tabular}[c]{@{}c@{}}~~Square~~ \\ (12/255)\end{tabular}}}  & \multirow{2}{*}{\textbf{\begin{tabular}[c]{@{}c@{}}~~GAMA~~ \\ (8/255)\end{tabular}}} \\
\\
\midrule
\multicolumn{1}{l}{}            & \multicolumn{4}{c}{\textbf{SVHN, PreActResNet18}}                                           \\
\midrule
AWP                             & 91.91             & 75.92                    & 35.78                      & \multicolumn{1}{c}{53.88}                 \\
Ours (tuned)                           & 94.61             & 78.37                    & 39.56                      & \multicolumn{1}{c}{55.15}                 \\
Ours (no tuning) & 95.17             & 78.16                    & 39.12                      & \multicolumn{1}{c}{54.77}                 \\
\midrule
\midrule
\multicolumn{1}{l}{\textbf{Method}}            & \textbf{Clean}    & \multirow{2}{*}{\textbf{\begin{tabular}[c]{@{}c@{}}~~GAMA~~ \\ (8/255)\end{tabular}}}    &  \multirow{2}{*}{\textbf{\begin{tabular}[c]{@{}c@{}}~~Square~~ \\ (16/255)\end{tabular}}}  &          \multirow{2}{*}{\textbf{\begin{tabular}[c]{@{}c@{}}~~GAMA~~ \\ (16/255)\end{tabular}}}                                 \\
\\
\midrule
\multicolumn{1}{l}{}            & \multicolumn{4}{c}{\textbf{CIFAR100, WRN-34-10}}                                             \\
\midrule
AWP                             & 62.73             & 29.92                    & 15.85                      & 11.55                                          \\
Ours (tuned)                           & 65.73             & 30.90                     & 18.47                      &  13.21                                         \\
Ours (no tuning) & 64.66             & 31.18                    & 17.93                      &  12.93   \\
\bottomrule
\end{tabular} %}
\vspace{-0.5cm}
\end{center}
\end{table}

% \vspace{0.7cm}
\noindent \textbf{Transferability of hyperparameters across datasets:} Although the proposed approach introduces two additional hyperparameters ($\alpha$ and $\lambda$), we show in Table-\ref{table:datasets_transfer} that even if we use the hyperparameters fine-tuned on CIFAR-10 dataset (WRN-34-10), they work well on SVHN (PreActResNet-18) and CIFAR-100 (WRN-34-10) datasets as well, thus showing good performance even without any fine-tuning. The gains obtained after fine-tuning specifically for the dataset are marginal. 
% \vspace{-0.4cm}

\end{document}